\documentclass[10pt,journal,compsoc]{IEEEtran}

\usepackage{color}
\usepackage{caption}
\usepackage{times}
\usepackage{helvet}
\usepackage{courier}
\usepackage{amsmath}
\usepackage{amssymb}
\usepackage{amsthm}
\usepackage{graphicx}
\usepackage{array}
\usepackage{booktabs}
\usepackage{amsopn}
\usepackage{amsmath,bm}
\usepackage{algorithm}
\usepackage{algorithmic}
\usepackage{enumerate}
\usepackage{color}
\usepackage{multirow}
\usepackage{bbding}
\usepackage{threeparttable}
\usepackage{threeparttable} 
\usepackage{dsfont} 
\usepackage{makecell} 
\usepackage{url}            

\newcommand{\eg}{\emph{e.g.}}
\newcommand{\etal}{\emph{et al.}}
\newcommand{\ie}{\emph{i.e.}}
\newcommand{\etc}{\emph{etc}}

\graphicspath{{figs/}}
\begin{document}

\title{A Survey on Visual Transformer}
\author{Kai Han, Yunhe Wang, Hanting Chen, Xinghao Chen, Jianyuan Guo, Zhenhua Liu, Yehui Tang, An Xiao, Chunjing Xu, Yixing Xu, Zhaohui Yang, Yiman Zhang, and Dacheng Tao~\IEEEmembership{Fellow,~IEEE}
\IEEEcompsocitemizethanks{
\IEEEcompsocthanksitem Kai Han, Hanting Chen, Xinghao Chen, Jianyuan Guo, Zhenhua Liu, Yehui Tang, An Xiao, Chunjing Xu, Yixing Xu, Zhaohui Yang, Yiman Zhang, and Yunhe Wang are with Huawei Noah's Ark Lab. E-mail: \{kai.han, yunhe.wang\}@huawei.com.
\IEEEcompsocthanksitem Hanting Chen, Zhenhua Liu, Yehui Tang, and Zhaohui Yang are also with School of EECS, Peking University.
\IEEEcompsocthanksitem Dacheng Tao is  with the School of Computer Science, in the Faculty of Engineering, at The University of Sydney, 6 Cleveland St, Darlington, NSW 2008, Australia. E-mail: dacheng.tao@sydney.edu.au.
\IEEEcompsocthanksitem Corresponding to Yunhe Wang and Dacheng Tao.
\IEEEcompsocthanksitem All authors are listed in alphabetical order of last name (except the primary and corresponding authors).
}
}
\markboth{A SUBMISSION TO IEEE TRANSACTION ON PATTERN ANALYSIS AND MACHINE INTELLIGENCE}
{Han \MakeLowercase{\textit{et al.}}: A Survey on Visual Transformer}

\IEEEtitleabstractindextext{%
\begin{abstract}
Transformer, first applied to the field of natural language processing, is a type of deep neural network mainly based on the self-attention mechanism. Thanks to its strong representation capabilities, researchers are looking at ways to apply transformer to computer vision tasks. In a variety of visual benchmarks, transformer-based models perform similar to or better than other types of networks such as convolutional and recurrent neural networks. Given its high performance and less need for vision-specific inductive bias, transformer is receiving more and more attention from the computer vision community. In this paper, we review these vision transformer models by categorizing them in different tasks and analyzing their advantages and disadvantages. The main categories we explore include the backbone network, high/mid-level vision, low-level vision, and video processing. We also include efficient transformer methods for pushing transformer into real device-based applications. Furthermore, we also take a brief look at the self-attention mechanism in computer vision, as it is the base component in transformer. Toward the end of this paper, we discuss the challenges and provide several further research directions for vision transformers.
\end{abstract}

}

\maketitle

\IEEEdisplaynontitleabstractindextext

\IEEEpeerreviewmaketitle

\IEEEraisesectionheading{\section{Introduction}\label{Sec:Introduction}}
\IEEEPARstart{D}{eep}  neural networks (DNNs) have become the fundamental infrastructure in today's artificial intelligence (AI) systems. Different types of tasks have typically involved different types of networks. For example, multi-layer perceptron (MLP) or the fully connected (FC) network is the classical type of neural network, which is composed of multiple linear layers and nonlinear activations stacked together~\cite{rosenblatt1957perceptron,rosenblatt1961principles}. Convolutional neural networks (CNNs) introduce convolutional layers and pooling layers for processing shift-invariant data such as images~\cite{lecun1998gradient,alexnet}. And recurrent neural networks (RNNs) utilize recurrent cells to process sequential data or time series data~\cite{rumelhart1985learning,LSTM}. Transformer is a new type of neural network. It mainly utilizes the self-attention mechanism~\cite{bahdanau2014neural,parikh2016decomposable} to extract intrinsic features~\cite{vaswani2017attention} and shows great potential for extensive use in AI applications.

\begin{figure*}[htp] 
	\centering
	\includegraphics[width=0.95\linewidth]{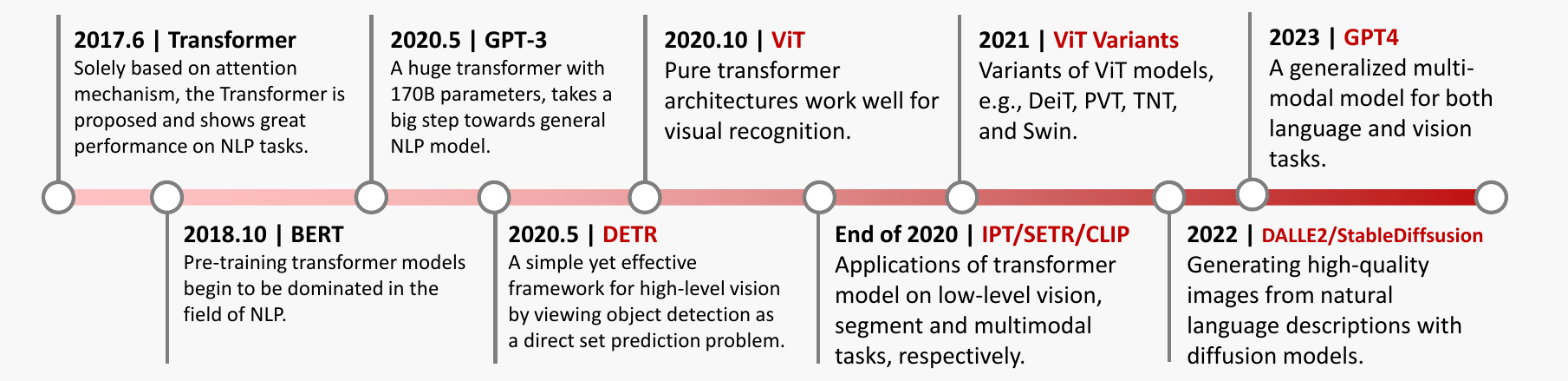}%
	\vspace{-0.5em}
	\caption{Key milestones in the development of transformer. The vision transformer models are marked in red.}
	\label{fig:timeline}
	\vspace{-1.5em}
\end{figure*}

Transformer was first applied to natural language processing (NLP) tasks where it achieved significant improvements~\cite{vaswani2017attention,bert,gpt3}. For example, Vaswani~\etal~\cite{vaswani2017attention} first proposed transformer based on attention mechanism for machine translation and English constituency parsing tasks. Devlin~\etal~\cite{bert} introduced a new language representation model called BERT (short for Bidirectional Encoder Representations from Transformers), which pre-trains a transformer on unlabeled text taking into account the context of each word as it is bidirectional. When BERT was published, it obtained state-of-the-art performance on 11 NLP tasks. Brown~\etal~\cite{gpt3} pre-trained a massive transformer-based model called GPT-3 (short for Generative Pre-trained Transformer 3) on 45 TB of compressed plaintext data using 175 billion parameters. It achieved strong performance on different types of downstream natural language tasks without requiring any fine-tuning. These transformer-based models, with their strong representation capacity, have achieved significant breakthroughs in NLP.

Inspired by the major success of transformer architectures in the field of NLP, researchers have recently applied transformer to computer vision (CV) tasks. In vision applications, CNNs are considered the fundamental component~\cite{resnet,fasterrcnn}, but nowadays transformer is showing it is a potential alternative to CNN. Chen~\etal~\cite{igpt} trained a sequence transformer to auto-regressively predict pixels, achieving results comparable to CNNs on image classification tasks. Another vision transformer model is ViT, which applies a pure transformer directly to sequences of image patches to classify the full image. Recently proposed by Dosovitskiy~\etal~\cite{vit}, it has achieved state-of-the-art performance on multiple image recognition benchmarks. In addition to image classification, transformer has been utilized to address a variety of other vision problems, including object detection~\cite{detr,ddetr}, semantic segmentation~\cite{SETR}, image processing~\cite{chen2020pre}, and video understanding~\cite{zhou2018end}. Thanks to its exceptional performance, more and more researchers are proposing transformer-based models for improving a wide range of visual tasks.

Due to the rapid increase in the number of transformer-based vision models, keeping pace with the rate of new progress is becoming increasingly difficult. As such, a survey of the existing works is urgent and would be beneficial for the community. In this paper, we focus on providing a comprehensive overview of the recent advances in vision transformers and discuss the potential directions for further improvement. To facilitate future research on different topics, we categorize the transformer models by their application scenarios, as listed in Table~\ref{tab:overview}. The main categories include backbone network, high/mid-level vision, low-level vision, and video processing. High-level vision deals with the interpretation and use of what is seen in the image~\cite{ullman1996high}, whereas mid-level vision deals with how this information is organized into what we experience as objects and surfaces~\cite{kimchi2003perceptual}. Given the gap between high- and mid-level vision is becoming more obscure in DNN-based vision systems~\cite{zhu2014top,fcn}, we treat them as a single category here. A few examples of transformer models that address these high/mid-level vision tasks include DETR~\cite{detr}, deformable DETR~\cite{ddetr} for object detection, and Max-DeepLab~\cite{wang2020max} for segmentation. Low-level image processing mainly deals with extracting descriptions from images (such descriptions are usually represented as images themselves)~\cite{fisher2008cvonline}. Typical applications of low-level image processing include super-resolution, image denoising, and style transfer. At present, only a few works~\cite{chen2020pre,parmar2018image} in low-level vision use transformers, creating the need for further investigation. Another category is video processing, which is an important part in both computer vision and image-based tasks. Due to the sequential property of video, transformer is inherently well suited for use on video tasks~\cite{zhou2018end,zeng2020learning}, in which it is beginning to perform on par with conventional CNNs and RNNs. Here, we survey the works associated with transformer-based visual models in order to track the progress in this field. Figure~\ref{fig:timeline} shows the development timeline of vision transformer — undoubtedly, there will be many more milestones in the future.

The rest of the paper is organized as follows. Section 2 discusses the formulation of the standard transformer and the self-attention mechanism. Section 4 is the main part of the paper, in which we summarize the vision transformer models on backbone, high/mid-level vision, low-level vision, and video tasks. We also briefly describe efficient transformer methods, as they are closely related to our main topic. In the final section, we give our conclusion and discuss several research directions and challenges. Due to the page limit, we describe the methods of transformer in NLP in the supplemental material, as the research experience may be beneficial for vision tasks. In the supplemental material, we also review the self-attention mechanism for CV as the supplementary of vision transformer models. In this survey, we mainly include the representative works (early, pioneering, novel, or inspiring works) since there are many preprinted works on arXiv and we cannot include them all in limited pages.

\begin{table*}[htb]
	\centering
	\renewcommand\arraystretch{1.0}
	\caption{Representative works of vision transformers.}
	\label{tab:overview}
	\vspace{-0.5em}
	\footnotesize
	\setlength{\tabcolsep}{3.5pt}{
		\begin{tabular}{c|c|c|c|c}
			\Xhline{1.2pt}
			Category & Sub-category & Method & Highlights & Publication \\
			\hline
			\multirow{5}{*}{\shortstack{Backbone}} & \multirow{3}{*}{Supervised pretraining}  & ViT~\cite{vit}   & Image patches, standard transformer & ICLR 2021  \\
			& & TNT~\cite{tnt}   & Transformer in transformer, local attention & NeurIPS 2021 \\
			& & Swin~\cite{cao2021swin}   & Shifted window, window-based self-attention & ICCV 2021 \\
			\cline{2-5}
			& \multirow{3}{*}{Self-supervised pretraining}  & iGPT~\cite{igpt}   & Pixel prediction self-supervised learning, GPT model & ICML 2020  \\
			& & MoCo v3~\cite{mocov3}   & Contrastive self-supervised learning, ViT & ICCV 2021 \\
			& & MAE~~\cite{mae} & Masked image modeling, ViT & CVPR 2022 \\
			\hline
			\multirow{9}{*}{\shortstack{High/Mid-level\\vision}} & \multirow{3}{*}{Object detection} & DETR~\cite{detr}   & Set-based prediction, bipartite matching, transformer  & ECCV 2020  \\
			& & Deformable DETR~\cite{ddetr}   & DETR, deformable attention module & ICLR 2021 \\
			& & UP-DETR~\cite{dai2020detr}   & Unsupervised pre-training, random query patch detection & CVPR 2021 \\
			\cline{2-5}
			& \multirow{3}{*}{Segmentation} & Max-DeepLab~\cite{wang2020max}   &PQ-style bipartite matching, dual-path transformer  & CVPR 2021  \\
			& & VisTR~\cite{wang2020end}   & Instance sequence matching and segmentation & CVPR 2021 \\
			& & SETR~\cite{SETR}   & Sequence-to-sequence prediction, standard transformer & CVPR 2021 \\
			\cline{2-5}
			& \multirow{3}{*}{Pose Estimation} & Hand-Transformer~\cite{huang2020hand}   & Non-autoregressive transformer, 3D point set  & ECCV 2020  \\
			& & HOT-Net~\cite{huang2020hot}   & Structured-reference extractor & MM 2020 \\
			& & METRO~\cite{lin2020end}   & Progressive dimensionality reduction & CVPR 2021 \\
			\hline
			\multirow{5}{*}{\shortstack{Low-level\\vision}} & \multirow{3}{*}{Image generation} & Image Transformer~\cite{parmar2018image} & Pixel generation using transformer  & ICML 2018  \\
			&  & Taming transformer~\cite{esser2021taming} & VQ-GAN, auto-regressive transformer  & CVPR 2021  \\
			&  & TransGAN~\cite{jiang2021transgan} & GAN using pure transformer architecture  & NeurIPS 2021  \\
			\cline{2-5}
			& \multirow{2}{*}{Image enhancement} & IPT~\cite{chen2020pre}   & Multi-task, ImageNet pre-training, transformer model & CVPR 2021  \\
			& & TTSR~\cite{yang2020learning}  & Texture transformer, RefSR & CVPR 2020 \\		
			
			\hline
			\multirow{2}{*}{\shortstack{Video\\processing}} & Video inpainting & STTN~\cite{zeng2020learning}  & Spatial-temporal adversarial loss  & ECCV 2020  \\
			\cline{2-5}
			& Video captioning &  Masked Transformer~\cite{zhou2018end}  & Masking network, event proposal  & CVPR 2018  \\
			\hline
			\multirow{4}{*}{\shortstack{Multimodality }} &\multirow{1}{*}{Classification}  & CLIP~\cite{clip}   & NLP supervision for images, zero-shot transfer & arXiv 2021 \\
			\cline{2-5}
			& \multirow{2}{*}{Image generation}  & DALL-E~\cite{dalle}   & Zero-shot text-to image generation & ICML 2021  \\
			&& Cogview~\cite{cogview} & VQ-VAE, Chinese input & NeurIPS 2021 \\
			\cline{2-5}
			& \multirow{1}{*}{Multi-task}  & GPT-4~\cite{gpt4}   & Large Multi-modal model for NLP \& CV tasks & arXiv 2023 \\
			\hline
			\multirow{4}{*}{\shortstack{Efficient\\transformer}} &\multirow{1}{*}{Decomposition}  & ASH~\cite{michel2019sixteen}   & Number of heads, importance estimation & NeurIPS 2019 \\
			\cline{2-5}
			& \multirow{1}{*}{Distillation}  & TinyBert~\cite{tinybert}   & Various losses for different modules & {\scriptsize EMNLP Findings 2020}  \\
			\cline{2-5}
			& \multirow{1}{*}{Quantization}  & FullyQT~\cite{prato2020fully}   & Fully quantized transformer & {\scriptsize EMNLP Findings 2020} \\
			\cline{2-5}
			& \multirow{1}{*}{Architecture design}  & ConvBert~\cite{jiang2020convbert}   & Local dependence, dynamic convolution & NeurIPS 2020 \\
			\hline
		\end{tabular}
	}
\vspace{-1em}
\end{table*}

\section{Formulation of Transformer}
\begin{figure}[htp] 
	\centering
	\includegraphics[width=0.55\columnwidth]{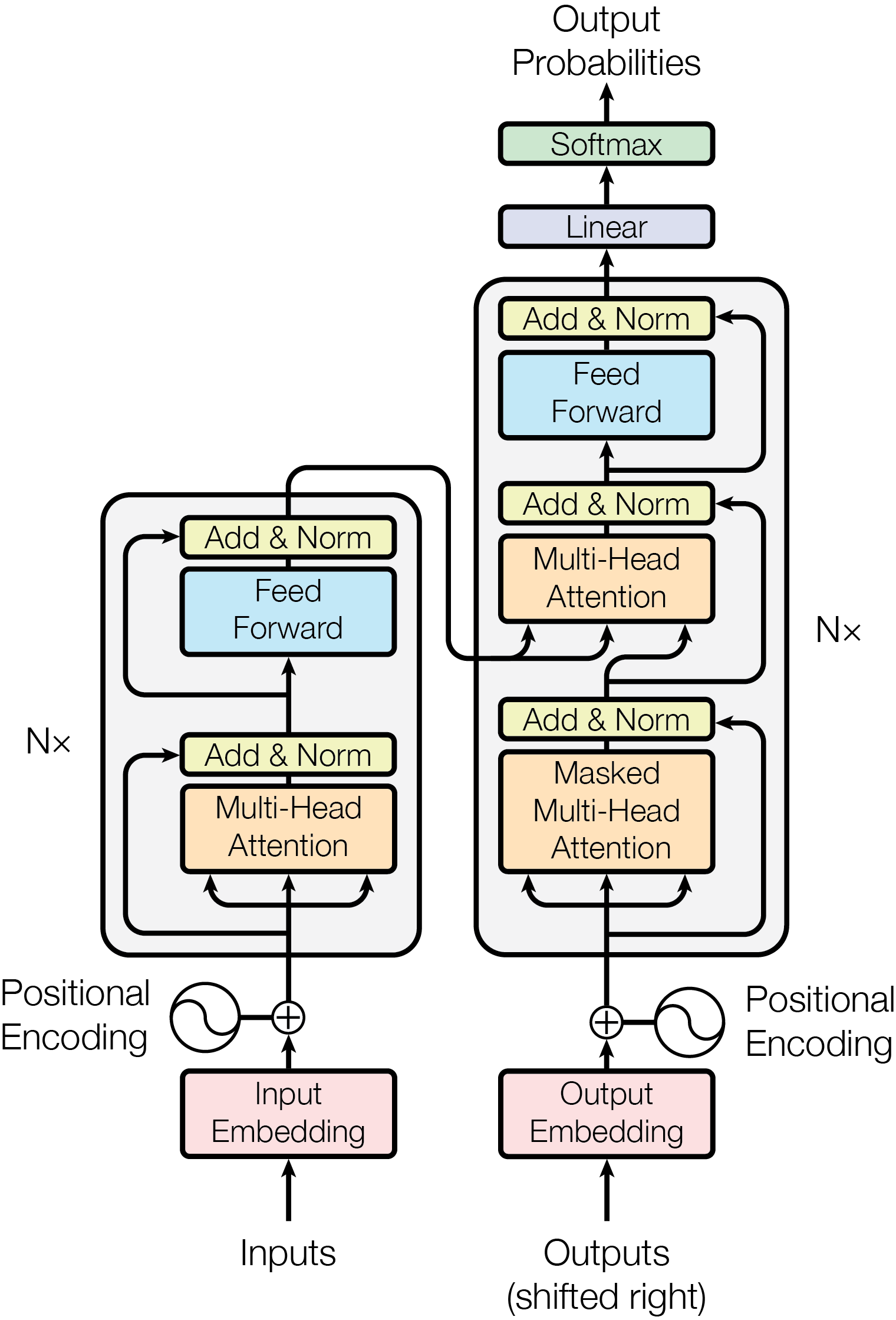}%
	\vspace{-0.5em}
	\caption{Structure of the original transformer (image from~\cite{vaswani2017attention}).}
	\label{fig:3-3}
	\vspace{-1em}
\end{figure}

Transformer~\cite{vaswani2017attention} was first used in the field of natural language processing (NLP) on machine translation tasks. As shown in Figure~\ref{fig:3-3}, it consists of an encoder and a decoder with several transformer blocks of the same architecture. The encoder generates encodings of inputs, while the decoder takes all the encodings and using their incorporated contextual information to generate the output sequence. Each transformer block is composed of a multi-head attention layer, a feed-forward neural network, shortcut connection and layer normalization. In the following, we describe each component of the transformer in detail.


\subsection{Self-Attention}
\begin{figure}[htp] 
	\centering
	\includegraphics[width=0.7\columnwidth]{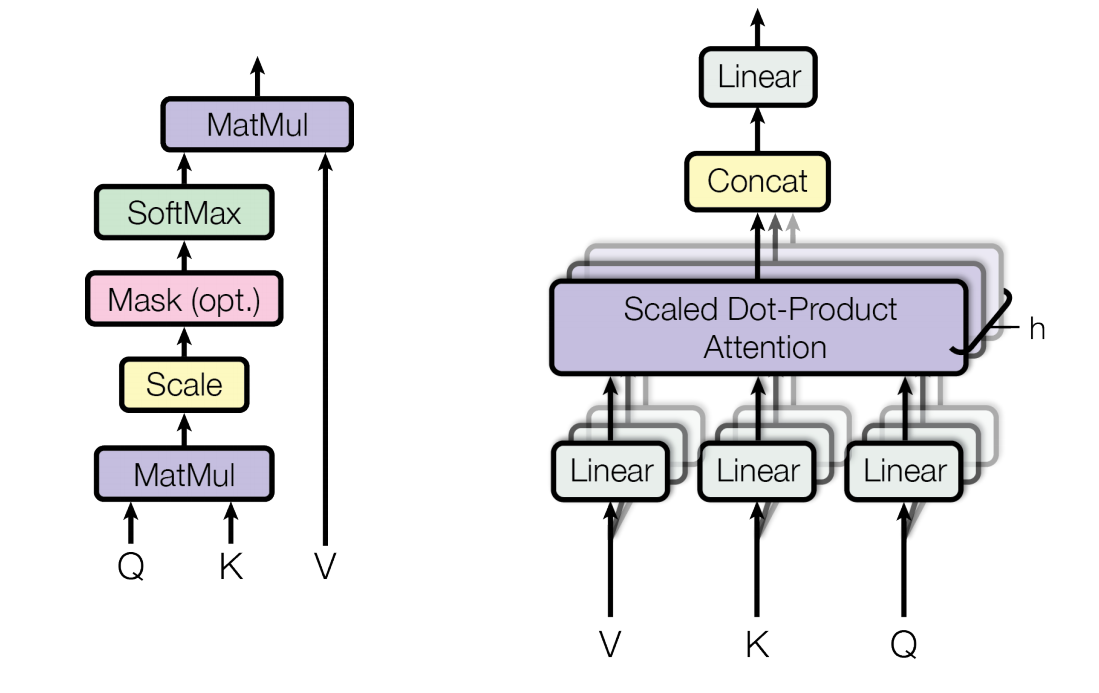}%
	\vspace{-1em}
	\caption{(Left) Self-attention process. (Right) Multi-head attention. The image is from~\cite{vaswani2017attention}.}
	\label{fig:3-2}
	\vspace{-1.5em}
\end{figure}

In the self-attention layer, the input vector is first transformed into three different vectors:  the query vector $\mathbf q$, the key vector $\mathbf k$ and the value vector $\mathbf v$ with dimension $d_q=d_k=d_v= d_{model}=512$. Vectors derived from different inputs are then packed together into three different matrices, namely, $\mathbf Q$, $\mathbf K$ and $\mathbf V$. Subsequently, the attention function between different input vectors is calculated as follows (and shown in Figure~\ref{fig:3-2} left):
\begin{itemize}
	\item \textbf{Step 1: }Compute scores between different input vectors with $\mathbf S=\mathbf Q\cdot \mathbf K^\top$;
	\item \textbf{Step 2: }Normalize the scores for the stability of gradient with $\mathbf S_n=\mathbf{S}/{\sqrt{d_k}}$;
	\item \textbf{Step 3: }Translate the scores into probabilities with softmax function $\mathbf P=\mathrm{softmax}(\mathbf S_n)$;
	\item \textbf{Step 4: }Obtain the weighted value matrix with $\mathbf Z=\mathbf V\cdot \mathbf P$.
\end{itemize}
The process can be unified into a single function:
\begin{equation}
\mathrm{Attention}(\mathbf Q,\mathbf K,\mathbf V)=\mathrm{softmax}(\frac{\mathbf Q\cdot \mathbf K^\top}{\sqrt{d_k}})\cdot \mathbf V.
\label{eq:3.1}
\end{equation}
The logic behind Eq.~\ref{eq:3.1} is simple. Step 1 computes scores between each pair of different vectors, and these scores determine the degree of attention that we give other words when encoding the word at the current position. Step 2 normalizes the scores to enhance gradient stability for improved training, and step 3 translates the scores into probabilities. Finally, each value vector is multiplied by the sum of the probabilities. Vectors with larger probabilities receive additional focus from the following layers.

The encoder-decoder attention layer in the decoder module is similar to the self-attention layer in the encoder module with the following exceptions: The key matrix $K$ and value matrix $V$ are derived from the encoder module, and the query matrix $Q$ is derived from the previous layer.

Note that the preceding process is invariant to the position of each word, meaning that the self-attention layer lacks the ability to capture the positional information of words in a sentence. However, the sequential nature of sentences in a language requires us to incorporate the positional information within our encoding. To address this issue and allow the final input vector of the word to be obtained, a positional encoding with dimension $d_{model}$ is added to the original input embedding. Specifically, the position is encoded with the following equations:
\begin{align}
PE(pos, 2i)=sin(\frac{pos}{10000^{\frac{2i}{d_{model}}}});\\
PE(pos, 2i+1)=cos(\frac{pos}{10000^{\frac{2i}{d_{model}}}}),
\end{align}
in which $pos$ denotes the position of the word in a sentence, and $i$ represents the current dimension of the positional encoding. In this way, each element of the positional encoding corresponds to a sinusoid, and it allows the transformer model to learn to attend by relative positions and extrapolate to longer sequence lengths during inference. In apart from the fixed positional encoding in the vanilla transformer, learned positional encoding~\cite{gehring2017convolutional} and relative positional encoding~\cite{shaw2018self} are also utilized in various models~\cite{bert,vit}.

\noindent\textbf{Multi-Head Attention.}
Multi-head attention is a mechanism that can be used to boost the performance of the vanilla self-attention layer. Note that for a given reference word, we often want to focus on several other words when going through the sentence. A single-head self-attention layer limits our ability to focus on one or more specific positions without influencing the attention on other equally important positions at the same time. This is achieved by giving attention layers different representation subspace. Specifically, different query, key and value matrices are used for different heads, and these matrices can project the input vectors into different representation subspace after training due to random initialization.

To elaborate on this in greater detail, given an input vector and the number of heads $h$, the input vector is first transformed into three different groups of vectors: the query group, the key group and the value group. In each group, there are $h$ vectors with dimension $d_{q'}=d_{k'}=d_{v'}=d_{model}/h=64$. The vectors derived from different inputs are then packed together into three different groups of matrices: $\{\mathbf Q_i\}_{i=1}^h$, $\{\mathbf K_i\}_{i=1}^h$ and $\{\mathbf V_i\}_{i=1}^h$. The multi-head attention process is shown as follows:
\begin{align}
\mathrm{MultiHead}(\mathbf Q', \mathbf K', \mathbf V') &=\mathrm{Concat}(\mathrm{head}_1, \cdots, \mathrm{head}_h)\mathbf W^o,\notag\\
\text{where } \mathrm{head}_i &= \mathrm{Attention}(\mathbf Q_i,\mathbf K_i,\mathbf V_i).
\end{align} 
Here, $\mathbf Q'$ (and similarly $\mathbf K'$ and $\mathbf V'$) is the concatenation of $\{\mathbf Q_i\}_{i=1}^h$, and $\mathbf W^o\in\mathbb R^{d_{model}\times d_{model}}$ is the projection weight.

\subsection{Other Key Concepts in Transformer}

\noindent\textbf{Feed-Forward Network.} A feed-forward network (FFN) is applied after the self-attention layers in each encoder and decoder. It consists of two linear transformation layers and a nonlinear activation function within them, and can be denoted as the following function:
\begin{equation}
\mathrm{FFN}(\mathbf X)=\mathbf W_2\sigma(\mathbf W_1\mathbf X),
\end{equation}
where $\mathbf W_1$ and $\mathbf W_2$ are the two parameter matrices of the two linear transformation layers, and $\sigma$ represents the nonlinear activation function, such as GELU~\cite{gelu}. The dimensionality of the hidden layer is $d_h=2048$.

\noindent\textbf{Residual Connection in the Encoder and Decoder.} As shown in Figure~\ref{fig:3-3}, a residual connection is added to each sub-layer in the encoder and decoder. This strengthens the flow of information in order to achieve higher performance. A layer-normalization~\cite{ba2016layer} is followed after the residual connection. The output of these operations can be described as:
\begin{equation}
\mathrm{LayerNorm}(\mathbf X+\mathrm{Attention}(\mathbf X)).
\end{equation}
Here, $\mathbf X$ is used as the input of self-attention layer, and the query, key and value matrices $\mathbf Q, \mathbf K$ and $\mathbf V$ are all derived from the same input matrix $\mathbf X$. A variant pre-layer normalization (Pre-LN) is also widely-used~\cite{baevski2018adaptive,wang2019learning,vit}. Pre-LN inserts the layer normalization inside the residual connection and before multi-head attention or FFN.
For the normalization layer, there are several alternatives such as batch normalization~\cite{bn}. Batch normalization usually perform worse when applied on transformer as the feature values change acutely~\cite{shen2020powernorm}. Some other normalization algorithms~\cite{xu2019understanding,shen2020powernorm,bachlechner2020rezero} have been proposed to improve training of transformer.

\noindent\textbf{Final Layer in the Decoder.} The final layer in the decoder is used to turn the stack of vectors back into a word. This is achieved by a linear layer followed by a softmax layer. The linear layer projects the vector into a logits vector with $d_{word}$ dimensions, in which $d_{word}$ is the number of words in the vocabulary. The softmax layer is then used to transform the logits vector into probabilities.

When used for CV tasks, most transformers adopt the original transformer's encoder module. Such transformers can be treated as a new type of feature extractor. Compared with CNNs which focus only on local characteristics, transformer can capture long-distance characteristics, meaning that it can easily derive global information. And in contrast to RNNs, whose hidden state must be computed sequentially, transformer is more efficient because the output of the self-attention layer and the fully connected layers can be computed in parallel and easily accelerated. From this, we can conclude that further study into using transformer in computer vision as well as NLP would yield beneficial results.

\section{Vision Transformer}
In this section, we review the applications of transformer-based models in computer vision, including image classification, high/mid-level vision, low-level vision and video processing. We also briefly summarize the applications of the self-attention mechanism and model compression methods for efficient transformer.

\begin{figure}[htp]
	\centering
	\includegraphics[width=0.9\linewidth]{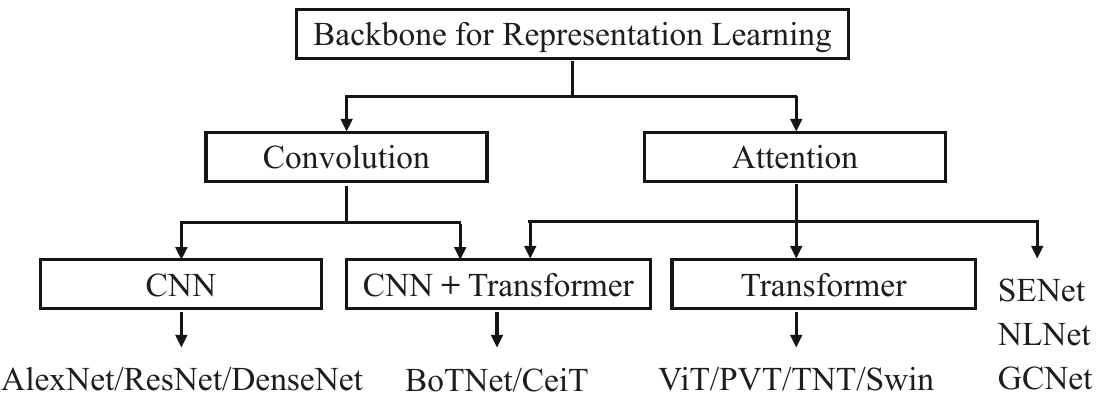}
	\caption{A taxonomy of backbone using convolution and attention.}
	\label{backbone-tree}
	\vspace{-1em}
\end{figure}

\subsection{Backbone for Representation Learning}
Inspired by the success that transformer has achieved in the field of NLP, some researchers have explored whether similar models can learn useful representations for images. Given that images involve more dimensions, noise and redundant modality compared to text, they are believed to be more difficult for generative modeling.

Other than CNNs, the transformer can be used as backbone networks for image classification. Wu~\etal~\cite{VT} adopted ResNet as a convenient baseline and used vision transformers to replace the last stage of convolutions. Specifically, they apply convolutional layers to extract low-level features that are then fed into the vision transformer. For the vision transformer, they use a \emph{tokenizer} to group pixels into a small number of \emph{visual tokens}, each representing a semantic concept in the image. These \emph{visual tokens} are used directly for image classification, with the transformers being used to model the relationships between tokens. As shown in Figure~\ref{backbone-tree}, the works can be divided into purely using transformer for vision and combining CNN and transformer. We summarize the results of these models in Table~\ref{table-vit} and Figure~\ref{acc-flops} to demonstrate the development of the backbones. In addition to supervised learning, self-supervised learning is also explored in vision transformer.

\subsubsection{Pure Transformer}
\begin{figure}[htp]
	\centering
	\includegraphics[width=1.0\linewidth]{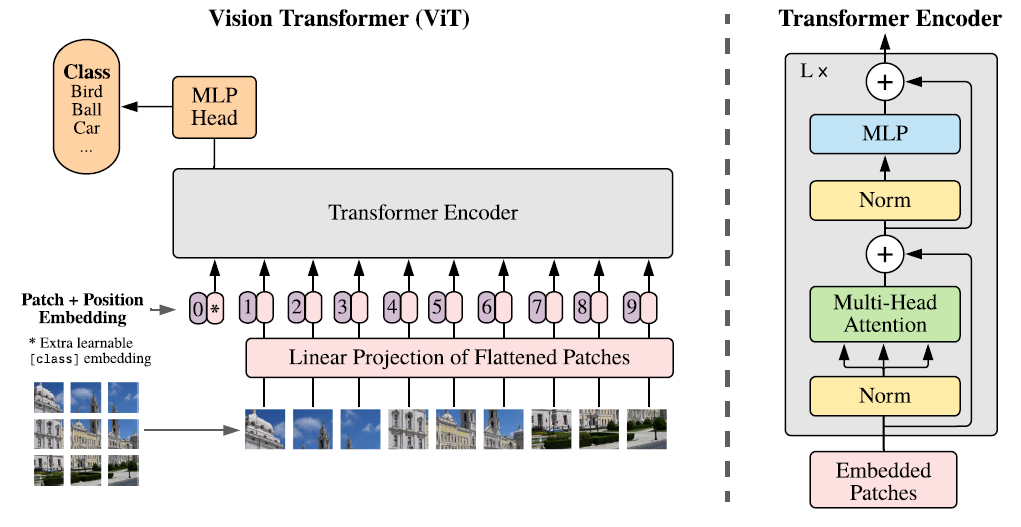}
	\vspace{-1em}
	\caption{The framework of ViT (image from~\cite{vit}).}
	\label{ViT}
	\vspace{-0.5em}
\end{figure}

\noindent\textbf{ViT.}
Vision Transformer (ViT)~\cite{vit} is a pure transformer directly applies to the sequences of image patches for image classification task. It follows transformer's original design as much as possible. Figure~\ref{ViT} shows the framework of ViT. 

To handle 2D images, the image $X\in \mathbb{R}^{h\times w \times c}$ is reshaped into a sequence of flattened 2D patches $X_p\in \mathbb{R}^{n\times (p^2 \cdot c)}$ such that $c$ is the number of channels. $(h,w)$ is the resolution of the original image, while $(p,p)$ is the resolution of each image patch. The effective sequence length for the transformer is therefore $n=hw/p^2$. Because the transformer uses constant widths in all of its layers, a trainable linear projection maps each vectorized path to the model dimension $d$, the output of which is referred to as patch embeddings. 

Similar to BERT's $[class]$ token, a learnable embedding is applied to the sequence of embedding patches. The state of this embedding serves as the image representation. During both pre-training and fine-tuning stage, the classification heads are attached to the same size. In addition, 1D position embeddings are added to the patch embeddings in order to retain positional information. It is worth noting that ViT utilizes only the standard transformer's encoder (except for the place for the layer normalization), whose output precedes an MLP head. In most cases, ViT is pre-trained on large datasets, and then fine-tuned for downstream tasks with smaller data.

ViT yields modest results when trained on mid-sized datasets such as ImageNet, achieving accuracies of a few percentage points below ResNets of comparable size. Because transformers lack some inductive biases inherent to CNNs--such as translation equivariance and locality--they do not generalize well when trained on insufficient amounts of data. However, the authors found that training the models on large datasets (14 million to 300 million images) surpassed inductive bias. When pre-trained  at sufficient scale, transformers achieve excellent results on tasks with fewer datapoints. For example, when pre-trained on the JFT-300M dataset, ViT approached or even exceeded state of the art performance on multiple image recognition benchmarks. Specifically, it reached an accuracy of 88.36\% on ImageNet, and 77.16\% on the VTAB suite of 19 tasks.

Touvron~\etal~\cite{DeiT} proposed a competitive convolution-free transformer, called Data-efficient image transformer (DeiT), by training on only the ImageNet database. DeiT-B, the reference vision transformer, has the same architecture as ViT-B and employs 86 million parameters. With a strong data augmentation, DeiT-B achieves top-1 accuracy of 83.1\% (single-crop evaluation) on ImageNet with no external data. In addition, the authors observe that using a CNN teacher gives better performance than using a transformer. Specifically, DeiT-B can achieve top-1 accuracy 84.40\% with the help of a token-based distillation. 

\noindent\textbf{Variants of ViT.}
Following the paradigm of ViT, a series of variants of ViT have been proposed to improve the performance on vision tasks. The main approaches include enhancing locality, self-attention improvement and architecture design.

The original vision transformer is good at capturing long-range dependencies between patches, but disregard the local feature extraction as the 2D patch is projected to a vector with simple linear layer. Recently, the researchers begin to pay attention to improve the modeling capacity for local information~\cite{tnt,liu2021swin,chen2021regionvit}. TNT~\cite{tnt} further divides the patch into a number of sub-patches and introduces a novel transformer-in-transformer architecture which utilizes an inner transformer block to model the relationship between sub-patches and an outer transformer block for patch-level information exchange. Twins~\cite{chu2021twins} and CAT~\cite{lin2021cat} alternately perform local and global attention layer-by-layer. Swin Transformers~\cite{liu2021swin,dong2021cswin} performs local attention within a window and introduces a shifted window partitioning approach for cross-window connections. Shuffle Transformer~\cite{huang2021shuffle,fang2021msg} further utilizes the spatial shuffle operation instead of shifted window partitioning to allow cross-window connections. RegionViT~\cite{chen2021regionvit} generates regional tokens and local tokens from an image, and local tokens receive global information via attention with regional tokens. In addition to the local attention, some other works propose to boost local information through local feature aggregation, \eg, T2T~\cite{yuan2021tokens}. These works demonstrate the benefit of the local information exchange and global information exchange in vision transformer.

As a key component of transformer, self-attention layer provides the ability for global interaction between image patches. Improving the calculation of self-attention layer has attracted many researchers. DeepViT~\cite{zhou2021deepvit} proposes to establish cross-head communication to re-generate the attention maps to increase the diversity at different layers. 
KVT~\cite{wang2021kvt} introduces the $k$-NN attention to utilize locality of images patches and ignore noisy tokens by only computing attentions with top-$k$ similar tokens. 
Refiner~\cite{zhou2021refiner} explores attention expansion in higher-dimension space and applied convolution to augment local patterns of the attention maps.
XCiT~\cite{el2021xcit} performs self-attention calculation across feature channels rather than tokens, which allows efficient processing of high-resolution images. The computation complexity and attention precision of the self-attention mechanism are two key-points for future optimization.

The network architecture is an important factor as demonstrated in the field of CNNs. The original architecture of ViT is a simple stack of the same-shape transformer block.
New architecture design for vision transformer has been an interesting topic. The pyramid-like architecture is utilized by many vision transformer models~\cite{wang2021pyramid,liu2021swin,sun2021visual,fan2021multiscale,zhang2021aggregating,pan2021less} including PVT~\cite{wang2021pyramid}, HVT~\cite{pan2021scalable}, Swin Transformer~\cite{liu2021swin} and PiT~\cite{heo2021rethinking}. There are also other types of architectures, such as two-stream architecture~\cite{chen2021crossvit} and U-net architecture~\cite{wang2021uformer,cao2021swin}. Neural architecture search (NAS) has also been investigated to search for better transformer architectures, \eg, Scaling-ViT~\cite{zhai2021scaling}, ViTAS~\cite{su2021vision}, AutoFormer~\cite{chen2021autoformer} and GLiT~\cite{chen2021glit}. Currently, both network design and NAS for vision transformer mainly draw on the experience of CNN. In the future, we expect the specific and novel architectures appear in the filed of vision transformer.

In addition to the aforementioned approaches, there are some other directions to further improve vision transformer, \eg, positional encoding~\cite{chu2021conditional,wu2021rethinking}, normalization strategy~\cite{touvron2021going}, shortcut connection~\cite{tang2021augmented} and removing attention~\cite{tolstikhin2021mlp,melas2021you,guo2021beyond,touvron2021resmlp}.

\begin{table}[]
	\centering
	\scriptsize
	\caption{\small ImageNet result comparison of representative CNN and vision transformer models. Pure transformer means only using a few convolutions in the stem stage. CNN + Transformer means using convolutions in the intermediate layers. Following~\cite{DeiT,liu2021swin}, the throughput is measured on NVIDIA V100 GPU and Pytorch, with 224$\times$224 input size.}
	\label{table-vit}
	\vspace{-1em}
	\begin{tabular}{l|c|c|c|c}
		\Xhline{1.2pt}
		\multirow{2}{*}{Model} & Params & FLOPs & Throughput & Top-1 \\ 
		& (M) & (B) & (image/s) & (\%) \\
		\hline
		\multicolumn{5}{c}{\textbf{CNN}} \\
		\hline
		ResNet-50~\cite{resnet,yuan2021tokens} & 25.6 & 4.1 & 1226 & 79.1 \\
		ResNet-101~\cite{resnet,yuan2021tokens} & 44.7 & 7.9  & 753 & 79.9  \\
		ResNet-152~\cite{resnet,yuan2021tokens} & 60.2 & 11.5  & 526 & 80.8  \\  
		\hline
		EfficientNet-B0~\cite{efficientnet} & 5.3 & 0.39  & 2694 & 77.1 \\
		EfficientNet-B1~\cite{efficientnet} & 7.8 & 0.70  & 1662 & 79.1 \\
		EfficientNet-B2~\cite{efficientnet} & 9.2 & 1.0  & 1255 & 80.1 \\
		EfficientNet-B3~\cite{efficientnet} & 12 & 1.8  & 732 & 81.6 \\
		EfficientNet-B4~\cite{efficientnet} & 19 & 4.2  & 349 & 82.9 \\
		\hline
		\multicolumn{5}{c}{\textbf{Pure Transformer}} \\
		\hline
		DeiT-Ti~\cite{vit,DeiT}  & 5 & 1.3   & 2536 & 72.2 \\
		DeiT-S~\cite{vit,DeiT}   & 22 & 4.6  & 940 & 79.8 \\
		DeiT-B~\cite{vit,DeiT}  & 86 & 17.6   & 292 & 81.8 \\  
		\hline
		
		T2T-ViT-14~\cite{yuan2021tokens}  & 21.5 & 5.2  & 764 & 81.5 \\
		T2T-ViT-19~\cite{yuan2021tokens}  & 39.2 & 8.9   & 464 & 81.9 \\
		T2T-ViT-24~\cite{yuan2021tokens}   & 64.1 & 14.1   & 312 & 82.3 \\  
		\hline
		
		PVT-Small~\cite{wang2021pyramid} & 24.5 & 3.8  & 820 & 79.8 \\ 
		PVT-Medium~\cite{wang2021pyramid} & 44.2 & 6.7  & 526 & 81.2 \\
		PVT-Large~\cite{wang2021pyramid} & 61.4 & 9.8  & 367 & 81.7 \\  
		\hline
		
		TNT-S~\cite{tnt} & 23.8 & 5.2  & 428 & 81.5  \\
		TNT-B~\cite{tnt} & 65.6 & 14.1  & 246 & 82.9  \\ 
		\hline
		
		CPVT-S~\cite{chu2021conditional} & 23 & 4.6  & 930 & 80.5 \\
		CPVT-B~\cite{chu2021conditional} & 88 & 17.6  & 285 & 82.3 \\  
		\hline		
		
		Swin-T~\cite{liu2021swin}  & 29 & 4.5  & 755 & 81.3
 \\ 
		Swin-S~\cite{liu2021swin}  & 50 & 8.7  &  437 & 83.0 \\
		Swin-B~\cite{liu2021swin}  & 88 & 15.4  &  278 & 83.3 \\ 
		\hline
		\multicolumn{5}{c}{\textbf{CNN + Transformer}} \\
		\hline
		Twins-SVT-S~\cite{chu2021twins} & 24 & 2.9  & 1059 & 81.7 \\
		Twins-SVT-B~\cite{chu2021twins} & 56 & 8.6  & 469  & 83.2 \\
		Twins-SVT-L~\cite{chu2021twins} & 99.2 & 15.1  & 288 & 83.7 \\
		\hline 
		
		Shuffle-T~\cite{huang2021shuffle}  & 29 & 4.6  & 791 & 82.5 \\
		Shuffle-S~\cite{huang2021shuffle}  & 50 & 8.9  &   450 & 83.5 \\ 
		Shuffle-B~\cite{huang2021shuffle}  &  88 & 15.6   &   279 & 84.0 \\	
		\hline
		CMT-S~\cite{guo2021cmt}  & 25.1 & 4.0  & 563 & 83.5 \\ 
		CMT-B~\cite{guo2021cmt}   & 45.7 & 9.3  &  285 & 84.5 \\ 
		\hline 
		
		VOLO-D1~\cite{yuan2021volo}  & 27 & 6.8  & 481 & 84.2 \\	
		VOLO-D2~\cite{yuan2021volo}  & 59 & 14.1 & 244 & 85.2 \\
		VOLO-D3~\cite{yuan2021volo}  & 86 & 20.6  & 168 & 85.4 \\
		VOLO-D4~\cite{yuan2021volo}  & 193 & 43.8   & 100 & 85.7 \\
		VOLO-D5~\cite{yuan2021volo}  & 296 & 69.0   & 64 & 86.1 \\
		
		\Xhline{1.2pt}
	\end{tabular}
\vspace{-1em}
\end{table}
\footnotetext[1]{}

\subsubsection{Transformer with Convolution}
Although vision transformers have been successfully applied to various visual tasks due to their ability to capture long-range dependencies within the input, there are still gaps in performance between transformers and existing CNNs. One main reason can be the lack of ability to extract local information. Except the above mentioned variants of ViT that enhance the locality, combining the transformer with convolution can be a more straightforward way to introduce the locality into the conventional transformer. 

There are plenty of works trying to augment a conventional transformer block or self-attention layer with convolution. For example, CPVT~\cite{chu2021conditional} proposed a conditional positional encoding (CPE) scheme, which is conditioned on the local neighborhood of input tokens and adaptable to arbitrary input sizes, to leverage convolutions for fine-level feature encoding. CvT~\cite{wu2021cvt}, CeiT~\cite{yuan2021incorporating}, LocalViT~\cite{li2021localvit} and CMT~\cite{guo2021cmt} analyzed the potential drawbacks when directly borrowing Transformer architectures from NLP and combined the convolutions with transformers together. Specifically, the feed-forward network (FFN) in each transformer block is combined with a convolutional layer that promotes the correlation among neighboring tokens. LeViT~\cite{graham2021levit} revisited principles from extensive literature on CNNs and applied them to transformers, proposing a hybrid neural network for fast inference image classification. BoTNet~\cite{srinivas2021bottleneck} replaced the spatial convolutions with global self-attention in the final three bottleneck blocks of a ResNet, and improved upon the baselines significantly on both instance segmentation and object detection tasks with minimal overhead in latency.

Besides, some researchers have demonstrated that transformer based models can be more difficult to enjoy a favorable ability of fitting data~\cite{vit,chen2021visformer,xiao2021early}, in other words, they are sensitive to the choice of optimizer, hyper-parameter, and the schedule of training. Visformer~\cite{chen2021visformer} revealed the gap between transformers and CNNs with two different training settings. The first one is the standard setting for CNNs, \ie, the training schedule is shorter and the data augmentation only contains random cropping and horizental flipping. The other one is the training setting used in~\cite{DeiT}, \ie, the training schedule is longer and the data augmentation is stronger. \cite{xiao2021early} changed the early visual processing of ViT by replacing its embedding stem with a standard convolutional stem, and found that this change allows ViT to converge faster and enables the use of either AdamW or SGD without a significant drop in accuracy. In addition to these two works, \cite{graham2021levit,guo2021cmt} also choose to add convolutional stem on the top of the transformer.

\begin{figure}[htp]
	\vspace{-0.5em}
	\centering
	\setlength{\tabcolsep}{2pt}{
		\begin{tabular}{cc}
			\makecell*[c]{\includegraphics[width=0.5\linewidth]{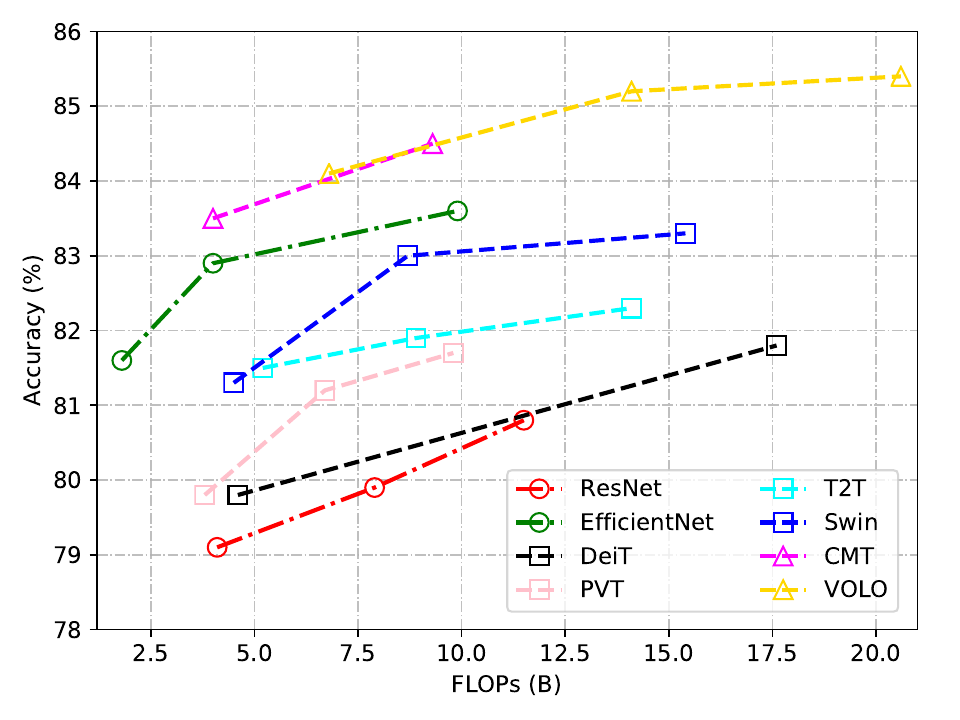}}  & \makecell*[c]{\includegraphics[width=0.5\linewidth]{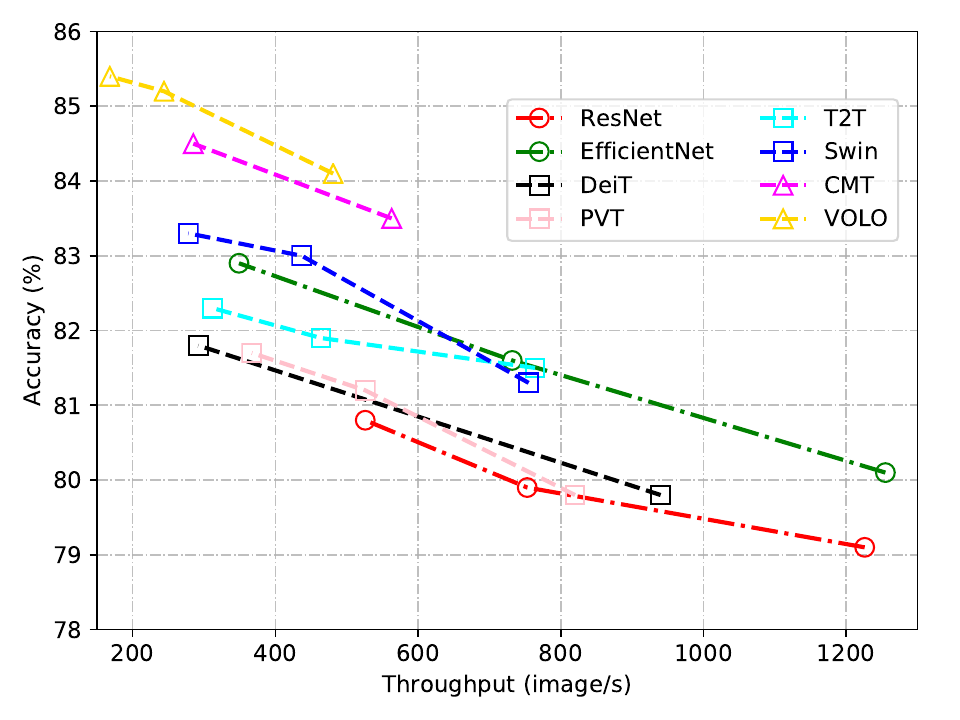}}
			\\
			\small (a) Acc v.s. FLOPs. & \small (b) Acc v.s. throughput.
		\end{tabular}
	}
	\vspace{-0.5em}
	\caption{FLOPs and throughput comparison of representative CNN and vision transformer models.}
	\label{acc-flops}
	\vspace{-1.0em}
\end{figure}

%

\subsubsection{Self-supervised Representation Learning}

\noindent\textbf{Generative Based Approach.}
Generative pre-training methods for images have existed for a long time~\cite{hinton1994autoencoders,vincent2008extracting,oord2016conditional,pathak2016context}. Chen~\etal~\cite{igpt} re-examined this class of methods and combined it with self-supervised methods. After that, several works~\cite{li2021mst,bao2021beit} were proposed to extend generative based self-supervised learning for vision transformer.

We briefly introduce iGPT~\cite{igpt} to demonstrate its mechanism. This approach consists of a pre-training stage followed by a fine-tuning stage. During the pre-training stage, auto-regressive and BERT objectives are explored. To implement pixel prediction, a sequence transformer architecture is adopted instead of language tokens (as used in NLP). Pre-training can be thought of as a favorable initialization or regularizer when used in combination with early stopping. During the fine-tuning stage, they add a small classification head to the model. This helps optimize a classification objective and adapts all weights.

The image pixels are transformed into a sequential data by $k$-means clustering. Given an unlabeled dataset ${X}$ consisting of high dimensional data $\mathbf x=(x_1,\cdots,x_n)$, they train the model by minimizing the negative log-likelihood
of the data:
\begin{equation}
	L_{AR} = \underset{\mathbf x\sim X}{\mathbb{E}} [-\log p(\mathbf x)],
\end{equation}
where $p(\mathbf x)$ is the probability density of the data of images, which can be modeled as:
\begin{equation}
	p(\mathbf x) = \prod_{i=1}^n p(x_{\pi_i}|x_{\pi_1},\cdots,x_{\pi_{i-1}},\theta).
\end{equation}
Here, the identity permutation $\pi_i=i$ is adopted for $1\leqslant i \leqslant n$, which is also known as raster order. Chen~\etal~also considered the BERT objective, which samples a sub-sequence $M\subset[1,n]$ such that each index $i$ independently has probability 0.15 of appearing in $M$. $M$ is called the BERT mask, and the model is trained by minimizing the negative log-likelihood of the ``masked'' elements $x_M$ conditioned on the ``unmasked'' ones $x_{[1,n] \backslash M}$:
\begin{equation}
	L_{BERT}=\underset{\mathbf x\sim X}{\mathbb{E}} \underset{M}{\mathbb{E}} \sum_{i\in M}[-\log p(x_i|x_{[1,n]\backslash M})].
\end{equation}
During the pre-training stage, they pick either $L_{AR}$ or $L_{BERT}$ and minimize the loss over the pre-training dataset. 

GPT-2~\cite{GPT2} formulation of the transformer decoder block is used. To ensure proper conditioning when training the AR objective, Chen~\etal~apply the standard upper triangular mask to the $n\times n$ matrix of attention logits. No attention logit masking is required when the BERT objective is used: Chen~\etal~zero out the positions after the content embeddings are applied to the input sequence. Following the final transformer layer, they apply a layer norm and learn a projection from the output to logits parameterizing the conditional distributions at each sequence element. When training BERT, they simply ignore the logits at unmasked positions. 


During the fine-tuning stage, they average pool the output of the final layer normalization layer across the sequence dimension to extract a $d$-dimensional vector of features per example.
They learn a projection from the pooled feature to class logits and use this projection to minimize a cross entropy loss. Practical applications offer empirical evidence that the joint objective of cross entropy loss and pretraining loss ($L_{AR}$ or $L_{BERT}$) works even better. After iGPT, masked image modeling is proposed such as MAE~\cite{mae} and SimMIM~\cite{simmim} which achieves competitive performance on downstream tasks.

iGPT and ViT are two pioneering works to apply transformer for visual tasks. The difference of iGPT and ViT-like models mainly lies on 3 aspects: 1) The input of iGPT is a sequence of color palettes by clustering pixels, while ViT uniformly divided the image into a number of local patches; 2) The architecture of iGPT is an encoder-decoder framework, while ViT only has transformer encoder; 3) iGPT utilizes auto-regressive self-supervised loss for training, while ViT is trained by supervised image classification task.

\noindent\textbf{Contrastive Learning Based Approach.}
Currently, contrastive learning is the most popular manner of self-supervised learning for computer vision. Contrastive learning has been applied on vision transformer for unsupervised pretraining~\cite{mocov3,xie2021self,li2021efficient}.

Chen~\etal~\cite{mocov3} investigate the effects of several fundamental components for training self-supervised ViT. The authors observe that instability is a major issue that degrades accuracy, and these results are indeed partial failure and they can be improved when training is made more stable. 

They introduce a ``MoCo v3'' framework, which is an incremental improvement of MoCo~\cite{mocov1}. Specifically, the authors take two crops for each image under random data augmentation. They are encodes by two encoders, $f_q$ and $f_k$, with output vectors $\mathbf q$ and $\mathbf k$. Intuitively, $\mathbf q$ behaves like a ``query'' and the goal of learning is to retrieve the corresponding ``key''. This is formulated as minimizing a contrastive loss function, which can be written as:
\begin{equation}
\mathcal{L}_q=-\text{log} \frac{\exp(\mathbf q \cdot \mathbf k^+ /\tau)}{\exp(\mathbf q \cdot \mathbf k^+ /\tau)+\sum_{\mathbf k^-} \exp(\mathbf q \cdot \mathbf k^- /\tau)}.
\end{equation}
Here $\mathbf k^+$ is $f_k$'s output on the same image as $\mathbf q$, known as $\mathbf q$'s positive sample. The set {$\mathbf k^-$} consists of $f_k$'s outputs from other images, known as $\mathbf q$'s negative samples. $\tau$ is a temperature hyper-parameter for $l_2$-normalized $\mathbf q$, $\mathbf k$. MoCo v3 uses the keys that naturally co-exist in the same batch and abandon the memory queue, which they find has diminishing gain if the batch is sufficiently large (\eg, 4096). With this simplification, the contrastive loss can be implemented in a simple way. The encoder $f_q$ consists of a backbone (\eg, ViT), a projection head and an extra prediction head; while the encoder $f_k$ has the backbone and projection head, but not the prediction head. $f_k$ is updated by the moving-average of $f_q$, excluding the prediction head.

MoCo v3 shows that the instability is a major issue of training the self-supervised ViT, thus they describe a simple trick that can improve the stability in various cases of the experiments. They observe that it is not necessary to train the patch projection layer. For the standard ViT patch size, the patch projection matrix is complete or over-complete. And in this case, random projection should be sufficient to preserve the information of the original patches. However, the trick alleviates the issue, but does not solve it. The model can still be unstable if the learning rate is too big and the first layer is unlikely the essential reason for the instability.

\subsubsection{Discussions}
All of the components of vision transformer including multi-head self-attention, multi-layer perceptron, shortcut connection, layer normalization, positional encoding and network topology, play key roles in visual recognition. As stated above, a number of works have been proposed to improve the effectiveness and efficiency of vision transformer. From the results in Figure~\ref{acc-flops}, we can see that combining CNN and transformer achieve the better performance, indicating their complementation to each other through local connection and global connection. Further investigation on backbone networks can lead to the improvement for the whole vision community. As for the self-supervised representation learning for vision transformer, we still need to make effort to pursue the success of large-scale pretraining in the filed of NLP.

\subsection{High/Mid-level Vision}
Recently there has been growing interest in using transformer for high/mid-level computer vision tasks, such as object detection~\cite{detr,ddetr,beal2020toward,yuan2020temporal,pointformer}, lane detection~\cite{liu2020end}, segmentation~\cite{wang2020end,wang2020max,SETR} and pose estimation~\cite{huang2020hand,huang2020hot,lin2020end,transpose}. We review these methods in this section.

\subsubsection{Generic Object Detection}

Traditional object detectors are mainly built upon CNNs, but transformer-based object detection has gained significant interest recently due to its advantageous capability. 

\begin{figure}[h]
	\begin{center}
		\includegraphics[width=\linewidth]{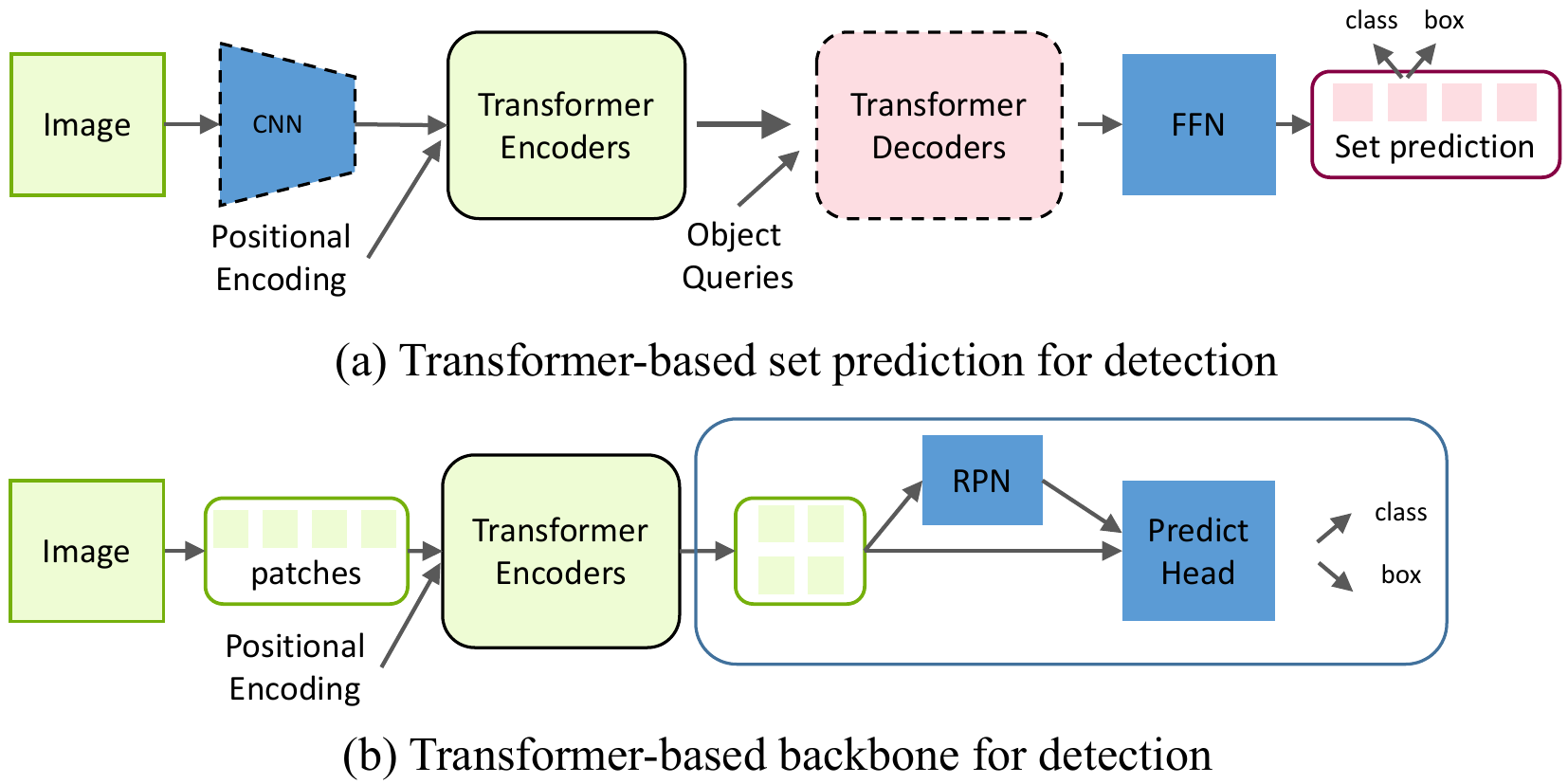}
	\end{center}
	\vspace{-1em}
	\caption{General framework of transformer-based object detection.}
	\label{fig:det_diagram}
	\vspace{-1.0em}
\end{figure}

Some object detection methods have attempted to use transformer's self-attention mechanism and then enhance the specific modules for modern detectors, such as feature fusion module~\cite{zhang2020feature} and prediction head~\cite{relationnet++}. We discuss this in the supplemental material.
%
%
Transformer-based object detection methods are broadly categorized into two groups: transformer-based set prediction methods~\cite{detr,ddetr,sun2020rethinking,zheng2020end,ma2021oriented} and transformer-based backbone methods~\cite{beal2020toward,pointformer}, as shown in Fig.~\ref{fig:det_diagram}. Transformer-based methods have shown strong performance compared with CNN-based detectors, in terms of both accuracy and running speed. Table~\ref{table:det_coco} shows the detection results for different transformer-based object detectors mentioned earlier on the COCO 2012 val set.

\begin{figure}[h]
	\begin{center}
		\includegraphics[width=\linewidth]{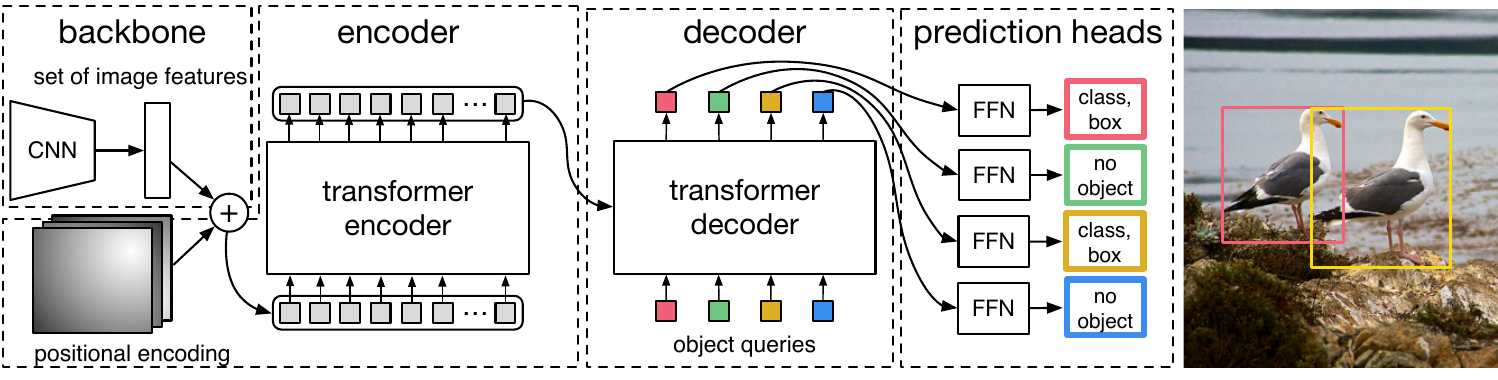}
	\end{center}
	\vspace{-0.5em}
	\caption{The overall architecture of DETR (image from ~\cite{detr}).}
	\label{fig:detr}
	\vspace{-1.0em}
\end{figure}

\noindent\textbf{Transformer-based Set Prediction for Detection}. As a pioneer for transformer-based detection method, the detection transformer (DETR) proposed by Carion~\etal~\cite{detr} redesigns the framework of object detection. DETR, a simple and fully end-to-end object detector, treats the object detection task as an intuitive set prediction problem, eliminating traditional hand-crafted components such as anchor generation and non-maximum suppression (NMS) post-processing. As shown in Fig.~\ref{fig:detr}, DETR starts with a CNN backbone to extract features from the input image. To supplement the image features with position information, fixed positional encodings are added to the flattened features before the features are fed into the encoder-decoder transformer. The decoder consumes the embeddings from the encoder along with $N$ learned positional encodings (object queries), and produces $N$ output embeddings. Here $N$ is a predefined parameter and typically larger than the number of objects in an image. Simple feed-forward networks (FFNs) are used to compute the final predictions, which include the bounding box coordinates and class labels to indicate the specific class of object (or to indicate that no object exists). Unlike the original transformer, which computes predictions sequentially, DETR decodes $N$ objects in parallel. DETR employs a bipartite matching algorithm to assign the predicted and ground-truth objects. As shown in Eq.~\ref{eq:hungarian_loss}, the Hungarian loss is exploited to compute the loss function for all matched pairs of objects.
\begin{equation}  \label{eq:hungarian_loss}
\scalebox{0.75}{$ \displaystyle
	{\cal L}_{\rm Hungarian}{(y, \hat{y})} = \sum_{i=1}^N \left[-\log  \hat{p}_{\hat{\sigma}(i)}(c_{i}) + {\mathds{1}_{\{c_i\neq\varnothing\}}} {\cal L}_{\rm box}{(b_{i}, \hat{b}_{\hat{\sigma}}(i))}\right]\,,
	$}
\end{equation}
where $\hat{\sigma}$ is the optimal assignment, $c_i$ and $\hat{p}_{\hat{\sigma}(i)}(c_{i})$ are the target class label and predicted label, respectively, and $b_i$ and $\hat{b}_{\hat{\sigma}}(i)$ are the ground truth and predicted bounding box, $y=\{(c_i, b_i)\}$ and $\hat y$ are the ground truth and prediction of objects, respectively.
DETR shows impressive performance on object detection, delivering comparable accuracy and speed with the popular and well-established Faster R-CNN~\cite{fasterrcnn} baseline on COCO benchmark.

DETR is a new design for the object detection framework based on transformer and empowers the community to develop fully end-to-end detectors. However, the vanilla DETR poses several challenges, specifically,  longer training schedule and poor performance for small objects.
To address these challenges, Zhu~\etal~\cite{ddetr} proposed Deformable DETR, which has become a popular method that significantly improves the detection performance. The deformable attention module attends to a small set of key positions around a reference point rather than looking at all spatial locations on image feature maps as performed by the original multi-head attention mechanism in transformer. This approach significantly reduces the computational complexity and brings benefits in terms of fast convergence. More importantly, the deformable attention module can be easily applied for fusing multi-scale features. Deformable DETR achieves better performance than DETR with $10\times$ less training cost and $1.6\times$ faster inference speed. And by using an iterative bounding box refinement method and two-stage scheme, Deformable DETR can further improve the detection performance.

There are also several methods to deal with the slow convergence problem of the original DETR. For example, Sun~\etal~\cite{sun2020rethinking} investigated why the DETR model has slow convergence and discovered that this is mainly due to the cross-attention module in the transformer decoder. To address this issue, an encoder-only version of DETR is proposed, achieving considerable improvement in terms of detection accuracy and training convergence. In addition, a new bipartite matching scheme is designed for greater training stability and faster convergence and two transformer-based set prediction models, \ie~TSP-FCOS and TSP-RCNN, are proposed to improve encoder-only DETR with feature pyramids. These new models achieve better performance compared with the original DETR model. Gao~\etal~\cite{gao2021fast} proposed the Spatially Modulated Co-Attention (SMCA) mechanism to accelerate the convergence by constraining co-attention responses to be high near initially estimated bounding box locations. By integrating the proposed SMCA module into DETR, similar mAP could be obtained with about 10$\times$ less training epochs under comparable inference cost.

Given the high computation complexity associated with DETR, Zheng~\etal~\cite{zheng2020end} proposed an Adaptive Clustering Transformer (ACT) to reduce the computation cost of pre-trained DETR. ACT adaptively clusters the query features using a locality sensitivity hashing (LSH) method and broadcasts the attention output to the queries represented by the selected prototypes. ACT is used to replace the self-attention module of the pre-trained DETR model without requiring any re-training. This approach significantly reduces the computational cost while the accuracy slides slightly. The performance drop can be further reduced by utilizing a multi-task knowledge distillation (MTKD) method, which exploits the original transformer to distill the ACT module with a few epochs of fine-tuning. Yao~\etal~~\cite{yao2021efficient} pointed out that the random initialization in DETR is the main reason for the requirement of multiple decoder layers and slow convergence. To this end, they proposed the Efficient DETR to incorporate the dense prior into the detection pipeline via an additional region proposal network. The better initialization enables them to use only one decoder layers instead of six layers to achieve competitive performance with a more compact network.

\begin{table*}[ht]
	\small
	\caption{\small Comparison of different transformer-based object detectors on COCO 2017 val set. Running speed (FPS) is evaluated on an NVIDIA Tesla V100 GPU as reported in~\cite{ddetr}. \textsuperscript\textdagger Estimated speed according to the reported number in the paper. \textsuperscript\textdaggerdbl ViT backbone is pre-trained on ImageNet-21k. $^*$ViT backbone is pre-trained on an private dataset with 1.3 billion images.}
	\label{table:det_coco}
	\vspace{-1.0em}
	\begin{center}
		\scriptsize
		\resizebox{0.85\linewidth}{!}{
			\begin{tabular}{l|c|cccccc|ccl}
				\Xhline{1.2pt}
				Method & Epochs & AP & AP$_\text{50}$ & AP$_\text{75}$ & AP$_\text{S}$ & AP$_\text{M}$ & AP$_\text{L}$ & \#Params (M) & GFLOPs & FPS\\
				\hline
				\textit{CNN based} &&&&&&&&&&\\
				FCOS~\cite{tian2019fcos} & 36 & 41.0 & 59.8 & 44.1 & 26.2 & 44.6 & 52.2 & - & 177 & 23\textsuperscript\textdagger\\
				Faster R-CNN + FPN~\cite{fasterrcnn} & 109 & 42.0 & 62.1 & 45.5 & 26.6 & 45.4 & 53.4  & 42 & 180 & 26 \\
				\hline
				\textit{CNN Backbone + Transformer Head} &&&&&&&&&&\\
				DETR~\cite{detr} & 500 & 42.0 & 62.4 & 44.2 & 20.5 & 45.8 & 61.1 & 41 & 86 & 28 \\
				DETR-DC5~\cite{detr} & 500 &  43.3 & 63.1 & 45.9 & 22.5 & 47.3 & 61.1 & 41 & 187 & 12 \\
				Deformable DETR~\cite{ddetr} & 50 & 46.2 & 65.2 & 50.0 & 28.8 & 49.2& 61.7& 40 & 173 & 19 \\
				TSP-FCOS~\cite{sun2020rethinking} & 36 & 43.1 & 62.3 & 47.0 & 26.6 & 46.8 & 55.9 & - & 189 & 20\textsuperscript\textdagger\\
				TSP-RCNN~\cite{sun2020rethinking} & 96 & {45.0} & {64.5} & {49.6} & {29.7} & {47.7} & 58.0 & - & 188 & 15\textsuperscript\textdagger\\
				ACT+MKKD (L=32)~\cite{zheng2020end} & - & 43.1 & - & - & 61.4 & 47.1 & 22.2 & - & 169 &  14\textsuperscript\textdagger \\
				SMCA~\cite{gao2021fast} & 108 & 45.6 & 65.5 & 49.1 & 25.9 & 49.3 & 62.6 & - & - & - \\ 
				Efficient DETR~\cite{yao2021efficient} & 36 & 45.1 & 63.1 & 49.1 & 28.3 & 48.4 & 59.0 & 35 & 210 & - \\ 
				{UP-DETR}~\cite{dai2020detr}  & 150 & {40.5} & 60.8 & 42.6  & 19.0 & {44.4} & {60.0} & 41 & -& -  \\
				{UP-DETR}~\cite{dai2020detr}  & 300 & {42.8} & 63.0 & 45.3& 20.8 & {47.1} & {61.7}  & 41 & -& - \\
				\hline
				\textit{Transformer Backbone + CNN Head} &&&&&&&&&&\\
				ViT-B/16-FRCNN\textsuperscript\textdaggerdbl
				~\cite{beal2020toward} & 21 & 36.6 & 56.3 & 39.3 & 17.4 & 40.0 & 55.5 &- & -& -  \\
				ViT-B/16-FRCNN$^*$~\cite{beal2020toward} & 21 & 37.8 &  57.4 & 40.1 & 17.8 & 41.4 & 57.3 & - & -& -  \\
				PVT-Small+RetinaNet~\cite{wang2021pyramid} & 12 & 40.4 & 61.3 & 43.0 & 25.0 & 42.9 & 55.7 & 34.2 & 118 & - \\ 
				Twins-SVT-S+RetinaNet~\cite{chu2021twins} & 12 & 43.0 & 64.2 & 46.3 & 28.0 & 46.4 & 57.5 & 34.3 & 104 & - \\ 
				Swin-T+RetinaNet~\cite{liu2021swin} & 12 & 41.5 & 62.1 & 44.2 & 25.1 & 44.9 & 55.5 & 38.5 & 118 & - \\ 
				Swin-T+ATSS~\cite{liu2021swin} & 36 & 47.2 & 66.5 & 51.3 & - & - & - & 36 & 215 & - \\ 
				\hline
				\textit{Pure Transformer based} &&&&&&&&&&\\
				PVT-Small+DETR~\cite{wang2021pyramid} & 50 & 34.7 & 55.7 & 35.4 & 12.0 & 36.4 & 56.7 & 40 & - & - \\ 
				TNT-S+DETR~\cite{tnt} & 50 & 38.2 & 58.9 & 39.4 & 15.5 & 41.1 & 58.8 & 39 & - & - \\ 
				YOLOS-Ti~\cite{fang2021you} & 300 & 30.0 & - & - & - & - & - & 6.5 & 21 & - \\ 
				YOLOS-S~\cite{fang2021you} & 150 & 37.6 & 57.6 & 39.2 & 15.9 & 40.2 & 57.3 & 28 & 179 & - \\ 
				YOLOS-B~\cite{fang2021you} & 150 & 42.0 & 62.2 & 44.5 & 19.5 & 45.3 & 62.1 & 127 & 537 & - \\ 
				\Xhline{2\arrayrulewidth}
			\end{tabular}
		}
	\end{center}
	\vspace{-1.5em}
\end{table*}

\noindent\textbf{Transformer-based Backbone for Detection}. Unlike DETR which redesigns object detection as a set prediction tasks via transformer, Beal~\etal~\cite{beal2020toward} proposed to utilize transformer as a backbone for common detection frameworks such as Faster R-CNN~\cite{fasterrcnn}. The input image is divided into several patches and fed into a vision transformer, whose output embedding features are reorganized according to spatial information before passing through a detection head for the final results. A massive pre-training transformer backbone could bring benefits to the proposed ViT-FRCNN. There are also quite a few methods to explore versatile vision transformer backbone design~\cite{tnt,wang2021pyramid,liu2021swin,chu2021twins} and transfer these backbones to traditional detection frameworks like RetinaNet~\cite{retinanet} and Cascade R-CNN~\cite{cai2018cascade}. For example, Swin Transformer~\cite{liu2021swin} obtains about 4 box AP gains over ResNet-50 backbone with similar FLOPs for various detection frameworks.

\noindent\textbf{Pre-training for Transformer-based Object Detection}.
Inspired by the pre-training transformer scheme in NLP, several methods have been proposed to explore different pre-training scheme for transformer-based object detection~\cite{dai2020detr,fang2021you,bar2021detreg}. Dai~\etal~\cite{dai2020detr} proposed unsupervised pre-training for object detection (UP-DETR). Specifically, a novel unsupervised pretext task named random query patch detection is proposed to pre-train the DETR model. With this unsupervised pre-training scheme, UP-DETR significantly improves the detection accuracy on a relatively small dataset (PASCAL VOC). On the COCO benchmark with sufficient training data, UP-DETR still outperforms DETR, demonstrating the effectiveness of the unsupervised pre-training scheme.

Fang~\etal~\cite{fang2021you} explored how to transfer the pure ViT structure that is pre-trained on ImageNet to the more challenging object detection task and proposed the YOLOS detector. To cope with the object detection task, the proposed YOLOS first drops the classification tokens in ViT and appends learnable detection tokens. Besides, the bipartite matching loss is utilized to perform set prediction for objects. With this simple pre-training scheme on ImageNet dataset, the proposed YOLOS shows competitive performance for object detection on COCO benchmark.

\subsubsection{Segmentation}
Segmentation is an important topic in computer vision community, which broadly includes panoptic segmentation, instance segmentation and semantic segmentation~\etc. Vision transformer has also shown impressive potential on the field of segmentation.

\noindent\textbf{Transformer for Panoptic Segmentation.} 
DETR~\cite{detr} can be naturally extended for panoptic segmentation tasks and achieve competitive results by appending a mask head on the decoder. Wang~\etal~\cite{wang2020max} proposed Max-DeepLab to directly predict panoptic segmentation results with a mask transformer, without involving surrogate sub-tasks such as box detection. Similar to DETR, Max-DeepLab streamlines the panoptic segmentation tasks in an end-to-end fashion and directly predicts a set of non-overlapping masks and corresponding labels. Model training is performed using a panoptic quality (PQ) style loss, but unlike prior methods that stack a transformer on top of a CNN backbone, Max-DeepLab adopts a dual-path framework that facilitates combining the CNN and transformer.

\noindent\textbf{Transformer for Instance Segmentation.}
VisTR, a transformer-based video instance segmentation model, was proposed by Wang~\etal~\cite{wang2020end} to produce instance prediction results from a sequence of input images. A strategy for matching instance sequence is proposed to assign the predictions with ground truths. In order to obtain the mask sequence for each instance, VisTR utilizes the instance sequence segmentation module to accumulate the mask features from multiple frames and segment the mask sequence with a 3D CNN.
Hu~\etal~\cite{hu2021istr} proposed an instance segmentation Transformer (ISTR) to predict low-dimensional mask embeddings, and match them with ground truth for the set loss. ISTR conducted detection and segmentation with a recurrent refinement strategy which is different from the existing top-down and bottom-up frameworks.
Yang~\etal~\cite{yang2021associating} investigated how to realize better and more efficient embedding learning to tackle the semi-supervised video object segmentation under challenging multi-object scenarios.
Some papers such as ~\cite{wu2021fully,dong2021solq} also discussed using Transformer to deal with segmentation task.

\noindent\textbf{Transformer for Semantic Segmentation.}
Zheng~\etal~\cite{SETR} proposed a transformer-based semantic segmentation network (SETR). SETR utilizes an encoder similar to ViT~\cite{vit} as the encoder to extract features from an input image. A multi-level feature aggregation module is adopted for performing pixel-wise segmentation.
Strudel~\etal~\cite{strudel2021segmenter} introduced \textit{Segmenter} which relies on the output embedding corresponding to image patches and obtains class labels with a point-wise linear decoder or a mask transformer decoder.
Xie~\etal~\cite{xie2021segformer} proposed a simple, efficient yet powerful semantic segmentation framework which unifies Transformers with lightweight multilayer perception (MLP) decoders, which outputs multiscale features and avoids complex decoders.

\noindent\textbf{Transformer for Medical Image Segmentation.}
Cao~\etal~\cite{cao2021swin} proposed an Unet-like pure Transformer for medical image segmentation, by feeding the tokenized image patches into the Transformer-based U-shaped Encoder-Decoder architecture with skip-connections for local-global semantic feature learning. Valanarasu~\etal~\cite{valanarasu2021medical} explored transformer-based solutions and study the feasibility of using transformer-based network architectures for medical image segmentation tasks and proposed a Gated Axial-Attention model which extends the existing architectures by introducing an additional control mechanism in the self-attention module. Cell-DETR~\cite{prangemeier2020attention}, based on the DETR panoptic segmentation model, is an attempt to use transformer for cell instance segmentation. It adds skip connections that bridge features between the backbone CNN and the CNN decoder in the segmentation head in order to enhance feature fusion. Cell-DETR achieves state-of-the-art performance for cell instance segmentation from microscopy imagery.

\subsubsection{Pose Estimation}

Human pose and hand pose estimation are foundational topics that have attracted significant interest from the research community. Articulated pose estimation is akin to a structured prediction task, aiming to predict the joint coordinates or mesh vertices from input RGB/D images. Here we discuss some methods~\cite{huang2020hand,huang2020hot,lin2020end,transpose} that explore how to utilize transformer for modeling the global structure information of human poses and hand poses.

\noindent\textbf{Transformer for Hand Pose Estimation.}
Huang~\etal~\cite{huang2020hand} proposed a transformer based network for 3D hand pose estimation from point sets. The encoder first utilizes a PointNet~\cite{qi2017pointnet} to extract point-wise features from input point clouds and then adopts standard multi-head self-attention module to produce embeddings. In order to expose more global pose-related information to the decoder, a feature extractor such as PointNet++~\cite{qi2017pointnet++} is used to extract hand joint-wise features, which are then fed into the decoder as positional encodings.
Similarly, Huang~\etal~\cite{huang2020hot} proposed HOT-Net (short for hand-object transformer network) for 3D hand-object pose estimation. Unlike the preceding method which employs transformer to directly predict 3D hand pose from input point clouds, HOT-Net uses a ResNet to generate initial 2D hand-object pose and then feeds it into a transformer to predict the 3D hand-object pose. A spectral graph convolution network is therefore used to extract input embeddings for the encoder.
Hampali~\etal~\cite{hampali2021handsformer} proposed to estimate the 3D poses of two hands given a single color image. Specifically, appearance and spatial encodings of a set of potential 2D locations for the joints of both hands were inputted to a transformer, and the attention mechanisms were used to sort out the correct configuration of the joints and outputted the 3D poses of both hands.

\noindent\textbf{Transformer for Human Pose Estimation.}
Lin~\etal~\cite{lin2020end} proposed a mesh transformer (METRO) for predicting 3D human pose and mesh from a single RGB image. METRO extracts image features via a CNN and then perform position encoding by concatenating a template human mesh to the image feature. A multi-layer transformer encoder with progressive dimensionality reduction is proposed to gradually reduce the embedding dimensions and finally produce 3D coordinates of human joint and mesh vertices. To encourage the learning of non-local relationships between human joints, METRO randomly mask some input queries during training.
Yang~\etal~\cite{transpose} constructed an explainable model named TransPose based on Transformer architecture and low-level convolutional blocks. The attention layers built in Transformer can capture long-range spatial relationships between keypoints and explain what dependencies the predicted keypoints locations highly rely on.
Li~\etal~\cite{li2021tokenpose} proposed a novel approach based on Token representation for human Pose estimation (TokenPose). Each keypoint was explicitly embedded as a token to simultaneously learn constraint relationships and appearance cues from images.
Mao~\etal~\cite{mao2021tfpose} proposed a human pose estimation framework that solved the task in the regression-based fashion. They formulated the pose estimation task into a sequence prediction problem and solve it by transformers, which bypass the drawbacks of the heatmap-based pose estimator.
Jiang~\etal~\cite{jiang2021skeletor} proposed a novel transformer based network that can learn a distribution over both pose and motion in an unsupervised fashion rather than tracking body parts and trying to temporally smooth them. The method overcame inaccuracies in detection and corrected partial or entire skeleton corruption.
Hao~\etal~\cite{hao2021test} proposed to personalize a human pose estimator given a set of test images of a person without using any manual annotations. The method adapted the pose estimator during test time to exploit person-specific information, and used a Transformer model to build a transformation between the self-supervised keypoints and the supervised keypoints.

\subsubsection{Other Tasks}

There are also quite a lot different high/mid-level vision tasks that have explored the usage of vision transformer for better performance. We briefly review several tasks below.

\noindent\textbf{Pedestrian Detection.}
Because the distribution of objects is very dense in occlusion and crowd scenes, 
additional analysis and adaptation are often required when common detection networks are applied to pedestrian detection tasks.
Lin~\etal~\cite{lin2020detr} revealed that sparse uniform queries and a weak attention field in the decoder result in performance degradation 
when directly applying DETR or Deformable DETR to pedestrian detection tasks. 
To alleviate these drawbacks, the authors proposes Pedestrian End-to-end Detector (PED), 
which employs a new decoder called Dense Queries and Rectified Attention field (DQRF) to support dense queries and alleviate the noisy or narrow attention field of the queries.
They also proposed V-Match, which achieves additional performance improvements by fully leveraging visible annotations.

\noindent\textbf{Lane Detection.}
Based on PolyLaneNet~\cite{tabelini2020polylanenet}, Liu~\etal~\cite{liu2020end} proposed a method called LSTR, 
which improves performance of curve lane detection by learning the global context with a transformer network. 
Similar to PolyLaneNet, LSTR regards lane detection as a task of fitting lanes with polynomials and uses neural networks to predict the parameters of polynomials. 
To capture slender structures for lanes and the global context, LSTR introduces a transformer network into the architecture.
This enables processing of low-level features extracted by CNNs. 
In addition, LSTR uses Hungarian loss to optimize network parameters. 
As demonstrated in~\cite{liu2020end}, LSTR outperforms PolyLaneNet, with 2.82\% higher accuracy and $3.65\times$ higher FPS using 5-times fewer parameters.
The combination of a transformer network, CNN and Hungarian Loss culminates in a lane detection framework that is precise, fast, and tiny. 
Considering that the entire lane line generally has an elongated shape and long-range,  
Liu~\etal~\cite{liu2021condlanenet} utilized a transformer encoder structure for more efficient context feature extraction. This transformer encoder structure improves the detection of the proposal points a lot,
which rely on contextual features and global information, especially in the case where the backbone network is a small model.

\noindent\textbf{Scene Graph.}
Scene graph is a structured representation of a scene that can clearly express the objects, attributes, and relationships between objects in the scene~\cite{Chang2021}. To generate scene graph, most of existing methods first extract image-based object representations and then do message propagation between them. Graph R-CNN~\cite{yang2018graph} utilizes self-attention to integrate contextual information from neighboring nodes in the graph. Recently, Sharifzadeh~\etal~\cite{Sharifzadeh2020} employed transformers over the extracted object embedding. Sharifzadeh~\etal~\cite{Sharifzadeh2021} proposed a new pipeline called \textit{Texema} and employed a pre-trained Text-to-Text Transfer Transformer (T5)~\cite{2020t5} to create structured graphs from textual input and utilized them to improve the relational reasoning module. The T5 model enables \textit{Texema} to utilize the knowledge in texts. 

\noindent\textbf{Tracking.}
Some researchers also explored to use transformer encoder-decoder architecture in template-based discriminative trackers, such as TMT~\cite{wang2021transformer}, TrTr~\cite{Zhao2021trtr} and TransT~\cite{Chen_2021_CVPR}. 
All these work use a Siamese-like tracking pipeline to do video object tracking and utilize the encoder-decoder network to replace explicit cross-correlation operation for global and rich contextual inter-dependencies.
Specifically, the transformer encoder and decoder are assigned to the template branch and the searching branch, respectively. 
In addition, Sun~\etal~proposed TransTrack~\cite{Sun2021TransTrack}, which is an online joint-detection-and-tracking pipeline. It utilizes the query-key mechanism to track pre-existing objects and introduces a set of learned object queries into the pipeline to detect new-coming objects.
The proposed TransTrack achieves 74.5\% and 64.5\% MOTA on MOT17 and MOT20 benchmark.

\noindent\textbf{Re-Identification.}
He~\etal~\cite{he2021transreid} proposed TransReID to investigate the application of pure transformers in the field of object re-identification (ReID).
While introducing transformer network into object ReID, TransReID slices with overlap to reserve local neighboring structures around the patches and introduces 2D bilinear interpolation to help handle any given input resolution.
With the transformer module and the loss function,  a strong baseline was proposed to achieve comparable performance with CNN-based frameworks.
Moreover, The jigsaw patch module (JPM) was designed to facilitate perturbation-invariant and robust feature representation of objects and the side information embeddings (SIE) was introduced  to encode side information.
The final framework TransReID achieves state-of-the-art performance on both person and vehicle ReID benchmarks.
Both Liu~\etal~\cite{Liu2021reid3view} and Zhang~\etal~\cite{Zhang2021stt} provided solutions for introducing transformer network into video-based person Re-ID.
And similarly, both of the them utilized separated transformer networks to refine spatial and temporal features, and then utilized a cross view transformer to aggregate multi-view features.

\noindent\textbf{Point Cloud Learning.}
A number of other works exploring transformer architecture for point cloud learning~\cite{engel2020point,guo2020pct,zhao2020point} have also emerged recently. For example, Guo~\etal~\cite{guo2020pct} proposed a novel framework that replaces the original self-attention module with a more suitable offset-attention module, which includes implicit Laplace operator and normalization refinement. In addition, Zhao~\etal~\cite{zhao2020point} designed a novel transformer architecture called Point Transformer. The proposed self-attention layer is invariant to the permutation of the point set, making it suitable for point set processing tasks. Point Transformer shows strong performance for semantic segmentation task from 3D point clouds. 

\subsubsection{Discussions}
As discussed in the preceding sections, transformers have shown strong performance on several high-level tasks, including detection, segmentation and pose estimation.
The key issues that need to be resolved before transformer can be adopted for high-level tasks relate to input embedding, position encoding, and prediction loss. Some methods propose improving the self-attention module from different perspectives, for example, deformable attention~\cite{ddetr}, adaptive clustering~\cite{zheng2020end} and point transformer~\cite{zhao2020point}. Nevertheless, exploration into the use of transformers for high-level vision tasks is still in the preliminary stages and so further research may prove beneficial. For example, is it necessary to use feature extraction modules such as CNN and PointNet before transformer for potential better performance? How can vision transformer be fully leveraged using large scale pre-training datasets as BERT and GPT-3 do in the NLP field? And is it possible to pre-train a single transformer model and fine-tune it for different downstream tasks with only a few epochs of fine-tuning? How to design more powerful architecture by incorporating prior knowledge of the specific tasks? Several prior works have performed preliminary discussions for the aforementioned topics and We hope more further research effort is conducted into exploring more powerful transformers for high-level vision.

\subsection{Low-level Vision}

Few works apply transformers on low-level vision fields, such as image super-resolution and generation. These tasks often take images as outputs (\emph{e.g.}, high-resolution or denoised images), which is more challenging than high-level vision tasks such as classification, segmentation, and detection, whose outputs are labels or boxes.

\begin{figure}[h]
	\vspace{-0.5em}
	\centering
	\includegraphics[width=0.65\linewidth]{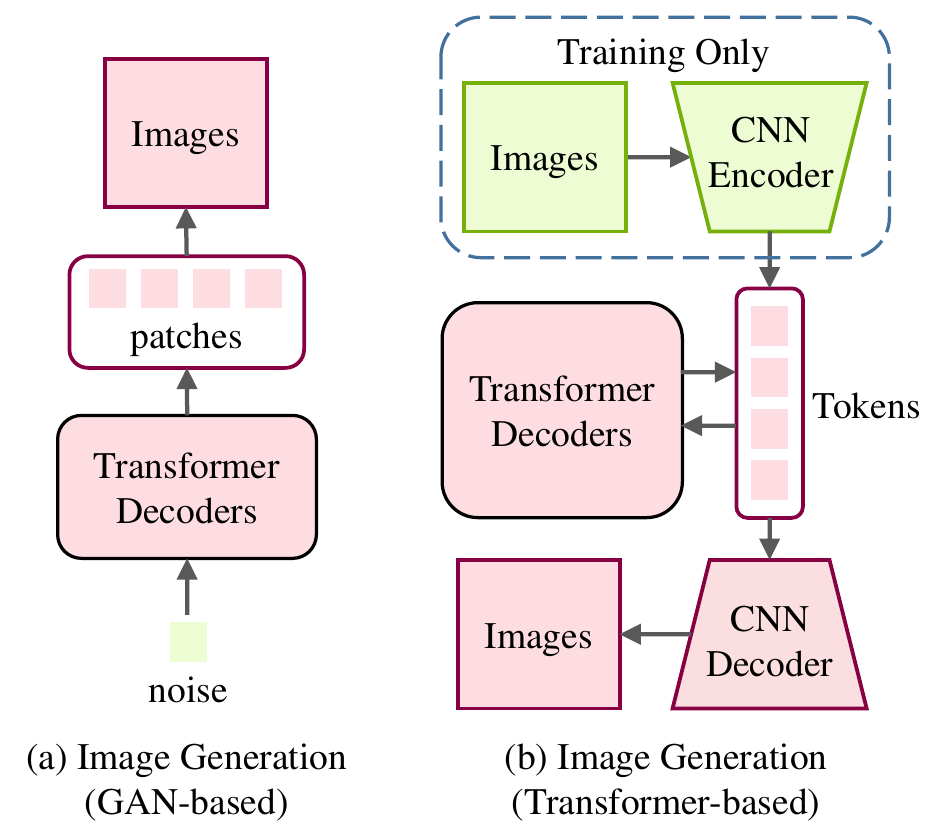}
	\vspace{-0.5em}
	\caption{A generic framework for transformer in image generation.}
	\label{fig:sec6-1}
	\vspace{-1.5em}
\end{figure}

\subsubsection{Image Generation}

An simple yet effective to apply transformer model to the image generation task is to directly change the architectures from CNNs to transformers, as shown in Figure~\ref{fig:sec6-1} (a). Jiang~\etal~\cite{jiang2021transgan} proposed TransGAN, which build GAN using the transformer architecture. Since the it is difficult to generate high-resolution images pixel-wise, a memory-friendly generator is utilized by gradually increasing the feature map resolution at different stages. Correspondingly, a multi-scale discriminator is designed to handle the varying size of inputs in different stages. Various training recipes are introduced including grid self-attention, data augmentation, relative position encoding and modified normalization to stabilize the training and improve its performance. Experiments on various benchmark datasets demonstrate the effectiveness and potential of the transformer-based GAN model in image generation tasks. Kwonjoon Lee~\etal~\cite{lee2021vitgan} proposed ViTGAN, which introduce several technique to both generator and discriminator to stabilize the training procedure and convergence. Euclidean distance is introduced for the self-attention module to enforce the Lipschitzness of transformer discriminator. Self-modulated layernorm and implicit neural representation are proposed to enhance the training for the generator. As a result, ViTGAN is the first work to demonstrate transformer-based GANs can achieve comparable performance to state-of-the-art CNN-based GANs.

Parmar~\etal~\cite{parmar2018image} proposed Image Transformer, taking the first step toward generalizing the transformer model to formulate image translation and generation tasks in an auto-regressive manner. Image Transformer consists of two parts: an encoder for extracting image representation and a decoder to generate pixels. For each pixel with value $0-255$, a $256 \times d$ dimensional embedding is learned for encoding each value into a $d$ dimensional vector, which is fed into the encoder as input. The encoder and decoder adopt the same architecture as that in~\cite{vaswani2017attention}. 
Each output pixel $q'$ is generated by calculating self-attention between the input pixel $q$ and previously generated pixels $m_1,m_2,...$ with position embedding $p_1,p_2,...$. For image-conditioned generation, such as super-resolution and inpainting, an encoder-decoder architecture is used, where the encoder's input is the low-resolution or corrupted images. For unconditional and class-conditional generation (\emph{i.e.}, noise to image), only the decoder is used for inputting noise vectors. Because the decoder's input is the previously generated pixels (involving high computation cost when producing high-resolution images), a local self-attention scheme is proposed. This scheme uses only the closest generated pixels as input for the decoder, enabling Image Transformer to achieve performance on par with CNN-based models for image generation and translation tasks, demonstrating the effectiveness of transformer-based models on low-level vision tasks.

Since it is difficult to directly generate high-resolution images by transformer models, Esser~\etal~\cite{esser2021taming} proposed Taming Transformer. Taming Transformer consists of two parts: a VQGAN and a transformer. VQGAN is a variant of VQVAE~\cite{oord2017neural}, which uses a discriminator and perceptual loss to improve the visual quality. Through VQGAN, the image can be represented by a series of context-rich discrete vectors and therefore these vectors can be easily predicted by a transformer model through an auto-regression way. The transformer model can learn the long-range interactions for generating high-resolution images. As a result, the proposed Taming Transformer achieves state-of-the-art results on a wide variety of image synthesis tasks.

Besides image generation, DALL$\cdot$E~\cite{dalle} proposed the transformer model for text-to-image generation, which synthesizes images according to the given captions. The whole framework consists of two stages. In the first stage, a discrete VAE is utilized to learn the visual codebook. In the second stage, the text is decoded by BPE-encode and the corresponding image is decoded by dVAE learned in the first stage. Then an auto-regression transformer is used to learn the prior between the encoded text and image. During the inference procedure, tokens of images are predicted by the transformer and decoded by the learned decoder. The CLIP model~\cite{clip} is introduced to rank generated samples. Experiments on text-to-image generation task demonstrate the powerful ability of the proposed model. Note that our survey mainly focus on pure vision tasks, we do not include the framework of DALL$\cdot$E in Figure~\ref{fig:sec6-1}. The image generation has been pushed to a higher level with the introduction of diffusion model~\cite{ddpm}, such as DALLE2~\cite{dalle2} and Stable Diffusion~\cite{rombach2022high}.

\subsubsection{Image Processing}

A number of recent works eschew using each pixel as the input for transformer models and instead use patches (set of pixels) as input. For example, Yang~\etal~\cite{yang2020learning} proposed Texture Transformer Network for Image Super-Resolution (TTSR), using the transformer architecture in the reference-based image super-resolution problem. It aims to transfer relevant textures from reference images to low-resolution images. Taking a low-resolution image and reference image as the query $\mathbf Q$ and key $\mathbf K$, respectively, the relevance $r_{i,j}$ is calculated between each patch $\mathbf q_i$ in $\mathbf Q$ and $\mathbf k_i$ in $\mathbf K$ as: 
\begin{equation}
	r_{i,j}=\left<\frac{\mathbf q_i}{\Vert \mathbf q_i\Vert},\frac{\mathbf k_i}{\Vert \mathbf k_i\Vert}\right>.
\end{equation} 
A hard-attention module is proposed to select high-resolution features $\mathbf V$ according to the reference image, so that the low-resolution image can be matched by using the relevance. The hard-attention map is calculated as:
\begin{equation}
	h_i = \arg \max_j r_{i,j}
\end{equation} 
The most relevant reference patch is $\mathbf t_i = \mathbf v_{h_i}$, where $\mathbf t_i$ in $\mathbf T$ is the transferred features. A soft-attention module is then used to transfer $\mathbf V$ to the low-resolution feature. The transferred features from the high-resolution texture image and the low-resolution feature are used to generate the output features of the low-resolution image. By leveraging the transformer-based architecture, TTSR can successfully transfer texture information from high-resolution reference images to low-resolution images in super-resolution tasks.

\begin{figure}[h]
	\centering
	\includegraphics[width=1.0\linewidth]{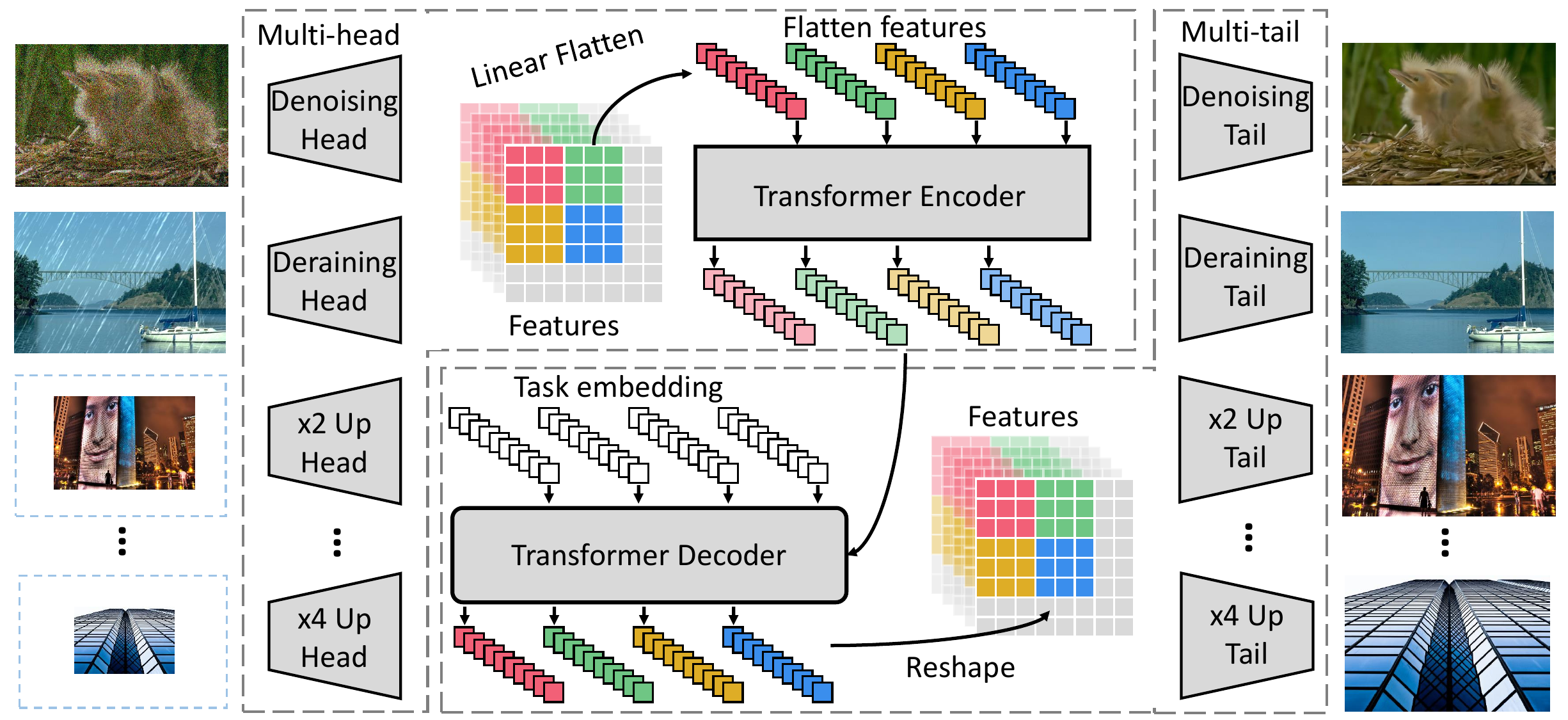}
	\caption{Diagram of IPT architecture (image from~\cite{chen2020pre}).}
	\label{fig:ipt}
	\vspace{-1.0em}
\end{figure}

Different from the preceding methods that use transformer models on single tasks, Chen~\etal~\cite{chen2020pre} proposed Image Processing Transformer (IPT), which fully utilizes the advantages of transformers by using large pre-training datasets. It achieves state-of-the-art performance in several image processing tasks, including super-resolution, denoising, and deraining. As shown in Figure~\ref{fig:ipt}, IPT consists of multiple heads, an encoder, a decoder, and multiple tails. The multi-head, multi-tail structure and task embeddings are introduced for different image processing tasks. The features are divided into patches, which are fed into the encoder-decoder architecture. Following this, the outputs are reshaped to features with the same size. Given the advantages of pre-training transformer models on large datasets, IPT uses the ImageNet dataset for pre-training. Specifically, images from this dataset are degraded by manually adding noise, rain streaks, or downsampling in order to generate corrupted images. The degraded images are used as inputs for IPT, while the original images are used as the optimization goal of the outputs. A self-supervised method is also introduced to enhance the generalization ability of the IPT model. Once the model is trained, it is fine-tuned on each task by using the corresponding head, tail, and task embedding. IPT largely achieves performance improvements on image processing tasks (\emph{e.g.}, 2 dB in image denoising tasks), demonstrating the huge potential of applying transformer-based models to the field of low-level vision.

Besides single image generation, Wang~\etal~\cite{wang2020sceneformer} proposed SceneFormer to utilize transformer in 3D indoor scene generation. By treating a scene as a sequence of objects, the transformer decoder can be used to predict series of objects and their location, category, and size. This has enabled SceneFormer to outperform conventional CNN-based methods in user studies.

\begin{figure}[h]
	\centering
	\includegraphics[width=0.5\linewidth]{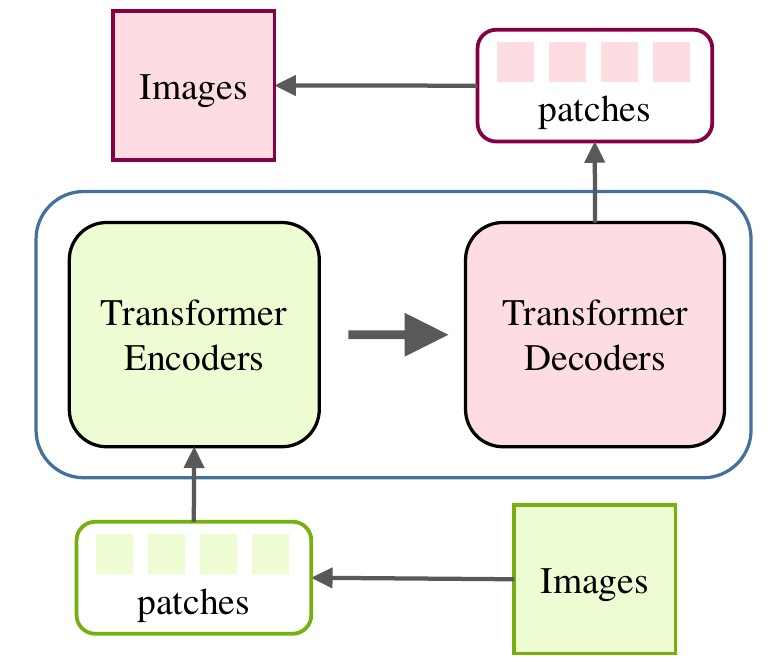}
	\vspace{-0.5em}
	\caption{A generic framework for transformer in image processing.}
	\label{fig:sec6-2}
	\vspace{-1.5em}
\end{figure}

It should be noted that iGPT~\cite{igpt} is pre-trained on an inpainting-like task. Since iGPT mainly focus on the fine-tuning performance on image classification tasks, we treat this work more like an attempt on image classification task using transformer than low-level vision tasks.

In conclusion, different to classification and detection tasks, the outputs of image generation and processing are images. Figure~\ref{fig:sec6-2} illustrates using transformers in low-level vision. In image processing tasks, the images are first encoded into a sequence of tokens or patches and the transformer encoder uses the sequence as input, allowing the transformer decoder to successfully produce desired images. In image generation tasks, the GAN-based models directly learn a decoder to generated patches to outputting images through linear projection, while the transformer-based models train a auto-encoder to learn a codebook for images and use an auto-regression transformer model to predict the encoded tokens. A meaningful direction for future research would be designing a suitable architecture for different image processing tasks.

\subsection{Video Processing}
Transformer performs surprisingly well on sequence-based tasks and especially on NLP tasks. In computer vision~(specifically, video tasks), spatial and temporal dimension information is favored, giving rise to the application of transformer in a number of video tasks, such as frame synthesis~\cite{liu2020convtransformer}, action recognition~\cite{girdhar2019video}, and video retrieval~\cite{liu2017two}.

\subsubsection{High-level Video Processing}

\noindent\textbf{Video Action Recognition.} Video human action tasks, as the name suggests, involves identifying and localizing human actions in videos. Context (such as other people and objects) plays a critical role in recognizing human actions. Rohit~\emph{et al.} proposed the action transformer~\cite{girdhar2019video} to model the underlying relationship between the human of interest and the surrounding context. Specifically, the I3D~\cite{carreira2017quo} is used as the backbone to extract high-level feature maps. The features extracted (using RoI pooling) from intermediate feature maps are viewed as the query (Q), while the key (K) and values (V) are calculated from the intermediate features. A self-attention mechanism is applied to the three components, and it outputs the classification and regressions predictions. Lohit~\emph{et al.}~\cite{lohit2019temporal} proposed an interpretable differentiable module, named temporal transformer network, to reduce the intra-class variance and increase the inter-class variance. In addition, Fayyaz and Gall proposed a temporal transformer~\cite{fayyaz2020sct} to perform action recognition tasks under weakly supervised settings. In addition to human action recognition, transformer has been utilized for group activity recognition~\cite{choi2009they}. Gavrilyuk~\emph{et al.} proposed an actor-transformer~\cite{gavrilyuk2020actor} architecture to learn the representation, using the static and dynamic representations generated by the 2D and 3D networks as input. The output of the transformer is the predicted activity.

\noindent\textbf{Video Retrieval.} The key to content-based video retrieval is to find the similarity between videos. Leveraging only the image-level of video-level features to overcome the associated challenges, Shao~\emph{et al.}~\cite{shaotemporal} suggested using the transformer to model the long-range semantic dependency. They also introduced the supervised contrastive learning strategy to perform hard negative mining. The results of using this approach on benchmark datasets demonstrate its performance and speed advantages. In addition, Gabeur~\emph{et al.}~\cite{gabeur2020multi} presented a multi-modal transformer to learn different cross-modal cues in order to represent videos.

\noindent\textbf{Video Object Detection.} To detect objects in a video, both global and local information is required. Chen~\emph{et al.} introduced the memory enhanced global-local aggregation~(MEGA)~\cite{chen2020memory} to capture more content. The representative features enhance the overall performance and address the \emph{ineffective} and \emph{insufficient} problems. Furthermore, Yin~\emph{et al.}~\cite{yin2020lidar} proposed a spatiotemporal transformer to aggregate spatial and temporal information. Together with another spatial feature encoding component, these two components perform well on 3D video object detection tasks.

\noindent\textbf{Multi-task Learning.} Untrimmed video usually contains many frames that are irrelevant to the target tasks. It is therefore crucial to mine the relevant information and discard the redundant information. To extract such information, Seong~\emph{et al.} proposed the video multi-task transformer network~\cite{seong2019video}, which handles multi-task learning on untrimmed videos. For the CoVieW dataset, the tasks are scene recognition, action recognition and importance score prediction. Two pre-trained networks on ImageNet and Places365 extract the scene features and object features. The multi-task transformers are stacked to implement feature fusion, leveraging the class conversion matrix~(CCM).

\subsubsection{Low-level Video Processing}

\noindent\textbf{Frame/Video Synthesis.} Frame synthesis tasks involve synthesizing the frames between two consecutive frames or after a frame sequence while video synthesis tasks involve synthesizing a video. Liu~\emph{et al.} proposed the ConvTransformer~\cite{liu2020convtransformer}, which is comprised of five components: feature embedding, position encoding, encoder, query decoder, and the synthesis feed-forward network. Compared with LSTM based works, the ConvTransformer achieves superior results with a more parallelizable architecture. Another transformer-based approach was proposed by Schatz~\emph{et al.}~\cite{schatz2020a}, which uses a recurrent transformer network to synthetize human actions from novel views.

\noindent\textbf{Video Inpainting.} Video inpainting tasks involve completing any missing regions within a frame. This is challenging, as it requires information along the spatial and temporal dimensions to be merged. Zeng~\emph{et al.} proposed a spatial-temporal transformer network~\cite{zeng2020learning}, which uses all the input frames as input and fills them in parallel. The spatial-temporal adversarial loss is used to optimize the transformer network.

\subsubsection{Discussions}
Compared to image, video has an extra dimension to encode the temporal information. Exploiting both spatial and temporal information helps to have a better understanding of a video. Thanks to the relationship modeling capability of transformer, video processing tasks have been improved by mining spatial and temporal information simultaneously. Nevertheless, due to the high complexity and much redundancy of video data, how to efficiently and accurately modeling both spatial and temporal relationships is still an open problem.

\subsection{Multi-Modal Tasks} 
Owing to the success of transformer across text-based NLP tasks, many researches are keen to exploit its potential for processing multi-modal tasks (\eg, video-text, image-text and audio-text). One example of this is VideoBERT~\cite{videobert}, which uses a CNN-based module to pre-process videos in order to obtain representation tokens. A transformer encoder is then trained on these tokens to learn the video-text representations for downstream tasks, such as video caption. Some other examples include VisualBERT~\cite{visualbert} and VL-BERT~\cite{vlbert}, which adopt a single-stream unified transformer to capture visual elements and image-text relationship for downstream tasks such as visual question answering (VQA) and visual commonsense reasoning (VCR). In addition, several studies such as SpeechBERT~\cite{speechbert} explore the possibility of encoding audio and text pairs with a transformer encoder to process auto-text tasks such as speech question answering (SQA).

\begin{figure}[htp]
	\centering
	\includegraphics[width=1.0\linewidth]{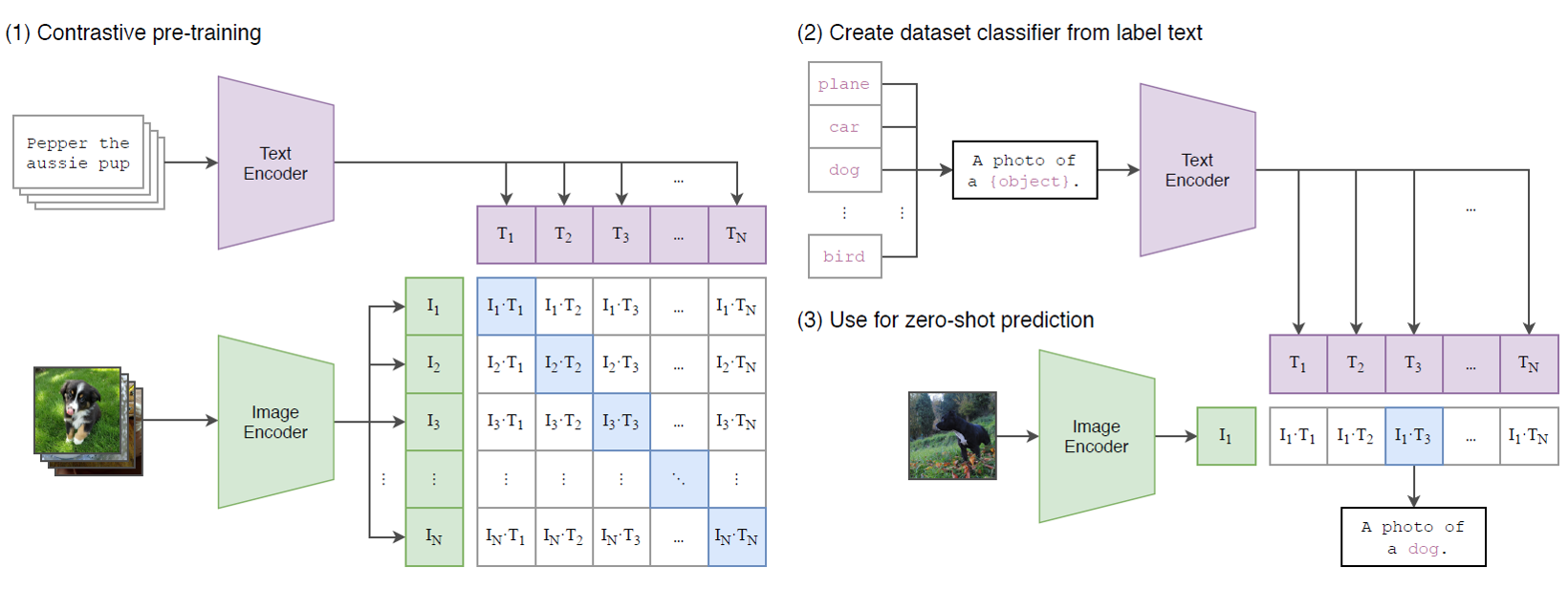}
	\caption{The framework of the CLIP (image from~\cite{clip}).}
	\label{CLIP}
	\vspace{-0.5em}
\end{figure}

Apart from the aforementioned pioneering multi-modal transformers, Contrastive Language-Image Pre-training (CLIP)~\cite{clip} takes natural language as supervision to learn more efficient image representation. CLIP jointly trains a text encoder and an image encoder to predict the corresponding training text-image pairs. The text encoder of CLIP is a standard transformer with masked self-attention used to preserve the initialization ability of the pre-trained language models. For the image encoder, CLIP considers two types of architecture, ResNet and Vision Transformer. CLIP is trained on a new dataset containing 400 million (image, text) pairs collected from the Internet. More specifically, given a batch of $N$ (image, text) pairs, CLIP learns both text and image embeddings jointly to maximize the cosine similarity of those $N$ matched embeddings while minimize $N^{2} - N$ incorrectly matched embeddings. On Zero-Shot transfer, CLIP demonstrates astonishing zero-shot classification performances, achieving $76.2\%$ top-1 accuracy on ImageNet-1K dataset without using any ImageNet training labels. Concretely, at inference, the text encoder of CLIP first computes the feature embeddings of all ImageNet Labels and the image encoder then computes the embeddings of all images. By calculating the cosine similarity of text and image embeddings, the text-image pair with the highest score should be the image and its corresponding label. Further experiments on 30 various CV benchmarks show the zero-shot transfer ability of CLIP and the feature diversity learned by CLIP.

While CLIP maps images according to the description in text, another work DALL-E~\cite{dalle} synthesizes new images of categories described in an input text. Similar to GPT-3, DALL-E is a multi-modal transformer with 12 billion model parameters autoregressively trained on a dataset of 3.3 million text-image pairs. More specifically, to train DALL-E, a two-stage training procedure is used, where in stage 1, a discrete variational autoencoder is used to compress 256$\times$ 256 RGB images into 32$\times$32 image tokens and then in stage 2, an autoregressive transformer is trained to model the joint distribution over the image and text tokens. Experimental results show that DALL-E can generate images of various styles from scratch, including photorealistic imagery, cartoons and emoji or extend an existing image while still matching the description in the text. Subsequently, Ding $\etal$ proposes CogView~\cite{cogview}, which is a transformer with VQ-VAE tokenizer similar to DALL-E, but supports Chinese text input. They claim CogView outperforms DALL-E and previous GAN-bsed methods and also unlike DALL-E, CogView does not need an additional CLIP model to rerank the samples drawn from transformer, $\ie$ DALL-E.

Recently, a Unified Transformer (UniT)~\cite{unit} model is proposed to cope with multi-modal multi-task learning, which can simultaneously handle multiple tasks across different domains, including object detection, natural language understanding and vision-language reasoning. Specifically, UniT has two transformer encoders to handle image and text inputs, respectively, and then the transformer decoder takes the single or concatenated encoder outputs according to the task modality. Finally, a task-specific prediction head is applied to the decoder outputs for different tasks. In the training stage, all tasks are jointly trained by randomly selecting a specific task within an iteration. The experiments show UniT achieves satisfactory performance on every task with a compact set of model parameters.

In conclusion, current transformer-based mutil-modal models demonstrates its architectural superiority for unifying data and tasks of various modalities, which demonstrates the potential of transformer to build a general-purpose intelligence agents able to cope with vast amount of applications. Future researches can be conducted in exploring the effective training or the extendability of multi-modal transformers (\eg, GPT-4~\cite{gpt4}).

\subsection{Efficient Transformer}\label{Sec:efficient}

Although transformer models have achieved success in various tasks, their high requirements for memory and computing resources block their implementation on resource-limited devices such as mobile phones. In this section, we review the researches carried out into compressing and accelerating transformer models for efficient implementation. This includes including network pruning, low-rank decomposition, knowledge distillation, network quantization, and compact architecture design. Table~\ref{tbl:compressed_tf]} lists some representative works for compressing transformer-based models.

\subsubsection{Pruning and Decomposition}

In transformer based pre-trained models (\eg, BERT), multiple  attention operations are performed in parallel to independently model the relationship between  different tokens~\cite{vaswani2017attention,bert}. However, specific tasks do not require all heads to be used.  For example, Michel~\etal~\cite{michel2019sixteen}  presented empirical evidence that a large percentage of attention heads can be removed at test time without  impacting performance significantly. The  number of heads required varies across different layers — some layers may even require only one head.  Considering the redundancy on attention heads, importance scores are defined to estimate the influence of each head on the final output in \cite{michel2019sixteen}, and unimportant heads can be removed for efficient deployment. Dalvi~\etal~\cite{prasanna2020bert}  analyzed the redundancy in pre-trained transformer models from two perspectives: general redundancy and task-specific redundancy. Following the lottery ticket hypothesis~\cite{frankle2018lottery}, Prasanna~\etal~\cite{prasanna2020bert} analyzed the lotteries in BERT and showed that good sub-networks also exist in transformer-based models, reducing both the FFN layers and attention heads in order to achieve high compression rates. For the vision transformer~\cite{vit} which splits an image to multiple patches, Tang~\etal~\cite{tang2021patch} proposed to reduce patch calculation to accelerate the inference, and the redundant patches can be automatically discovered by considering their contributions to the effective output features. Zhu~\etal~\cite{zhu2021visual} extended the network slimming approach~\cite{liu2017learning} to vision transformers for reducing the dimensions of linear projections in both FFN and attention modules.  

\begin{table}[b]	
	\vspace{-1em}
	\centering
	\renewcommand\arraystretch{1.0}
	\setlength{\tabcolsep}{2pt}{
		\begin{threeparttable}[b]
			\caption{List of representative compressed transformer-based models. The data of the Table is from~\cite{PTMS}.}
			\vspace{-0.5em}
			\footnotesize
			\begin{tabular}{c|c|c|c|c}
				\Xhline{1.2pt}
				Models & Compress Type & $\#$Layer &Params & Speed Up \\		
				\hline
				BERT$_{BASE}~\cite{bert}$& Baseline & 12 & 110M & $\times$1\\
				\hline
				ALBERT~\cite{albert} & Decomposition & 12 & 12M & $\times$5.6 \\
				\hline
				BERT-& Architecture & \multirow{2}*{6} & \multirow{2}*{66M} & \multirow{2}*{$\times$1.94}\\
				of-Theseus~\cite{bertofth}& design &  &  & \\
				\hline
				Q-BERT~\cite{QBERT} & \multirow{2}*{Quantization} & 12 & \multirow{2}*{-} & \multirow{2}*{-} \\
				Q8BERT~\cite{q8bert} & &12 & & \\
				\hline
				TinyBERT~\cite{tinybert} & \multirow{5}*{Distillation} & 4 & 14.5M & $\times$9.4 \\
				DistilBERT~\cite{distilbert} & & 6 & 6.6m & $\times$1.63 \\
				BERT-PKD~\cite{bertokd} & & 3$\sim$6 & 45.7$\sim$67M & $\times$3.73$\sim$1.64 \\
				MobileBERT~\cite{mobilebert} & &24 &25.3M & $\times$4.0 \\
				PD~\cite{pd} & & 6 &67.5M &$\times$2.0\\
				\hline
			\end{tabular}
			\label{tbl:compressed_tf]}
	\end{threeparttable}}
\end{table}

In addition to the width of transformer models, the depth (\ie, the number of layers) can also be reduced to accelerate the inference process~\cite{fan2019reducing,hou2020dynabert}.
Differing from the concept that different attention heads in transformer models can be computed in parallel, different layers have to be calculated sequentially because the input of the next layer depends on the output of previous layers. Fan~\etal~\cite{fan2019reducing} proposed a layer-wisely dropping strategy to regularize the training of models, and then the whole layers are removed together at the test phase. 

Beyond the pruning methods that directly discard modules in transformer models, matrix decomposition aims to approximate the large matrices with multiple small matrices based on the low-rank assumption. For example, Wang~\etal~\cite{wang2019structured} decomposed the standard matrix multiplication in transformer models, improving the inference efficiency. 

\subsubsection{Knowledge Distillation}

Knowledge distillation aims to train student networks by transferring knowledge from large teacher networks~\cite{hinton2015distilling,bucilua2006model,ba2014deep}. Compared  with teacher networks, student networks usually have thinner and shallower architectures, which  are easier to be deployed on resource-limited resources. 
Both the output and  intermediate features of  neural networks can also be used to transfer effective information from teachers  to students. Focused on  transformer models, Mukherjee~\etal~\cite{mukherjee2020xtremedistil} used the pre-trained BERT~\cite{bert} as a teacher to guide the training  of small models, leveraging large amounts of unlabeled data.  Wang~\etal~\cite{wang2020minilm}  train the student networks to mimic the output of self-attention layers in the pre-trained teacher models. The dot-product between values is introduced as a new form of knowledge for guiding  students. A teacher's assistant~\cite{mirzadeh2020improved} is also introduced in \cite{wang2020minilm}, reducing the gap between  large pre-trained transformer models and compact student networks, thereby facilitating the mimicking process. Due to the various types of layers in the transformer model (\ie, self-attention layer, embedding layer, and prediction layers),  Jiao~\etal~\cite{tinybert} design different objective functions to transfer knowledge from teachers to students. For example, the outputs of student models' embedding layers  imitate those of teachers via MSE losses.  For the vision transformer, Jia~\etal~\cite{jia2021efficient} proposed a fine-grained manifold distillation method, which excavates effective knowledge through the relationship between images and the
divided patches. 


\subsubsection{Quantization}

Quantization aims to reduce the number of bits needed to represent network weight or intermediate features~\cite{vanhoucke2011improving,yang2020searching}. 
Quantization methods for general neural networks have been discussed at length and achieve  performance on par with the original networks~\cite{park2020profit,fromm2020riptide,bai2018proxquant}. Recently, there has been growing interest in how to specially quantize transformer models~\cite{bhandare2019efficient,fanquantized}. For example, Shridhar~\etal~\cite{shridhar2020end} suggested embedding the input into binary high-dimensional vectors, and then using the binary input representation to train the binary neural networks. 
Cheong~\etal~\cite{cheong2019transformers} represented the weights in the transformer models by low-bit~(\eg, 4-bit) representation. 
Zhao~\etal~\cite{zhao2020investigation} empirically investigated various quantization methods and showed that k-means quantization has a huge development potential. Aimed at machine translation tasks, Prato~\etal~\cite{prato2020fully} proposed a fully quantized transformer, which, as the paper claims, is the first 8-bit model not to suffer any loss in translation quality. Beside, Liu~\etal~\cite{liu2021post} explored a post-training quantization scheme to reduce the memory storage and computational costs of vision transformers.

\subsubsection{Compact Architecture Design}
Beyond compressing  pre-defined transformer models into smaller ones, some works attempt to design compact models directly~\cite{wu2020lite,jiang2020convbert}. Jiang~\etal~\cite{jiang2020convbert} simplified the calculation of self-attention by proposing a new module — called span-based dynamic convolution — that combine the fully-connected layers and the convolutional layers. Interesting ``hamburger'' layers are proposed in \cite{anonymous2021is}, using matrix decomposition to substitute the original self-attention layers.Compared with standard self-attention operations, matrix decomposition can be calculated more efficiently while clearly reflecting the dependence between different tokens. The design of efficient transformer architectures can also be automated with neural architecture search~(NAS)~\cite{guo2019nat,so2019evolved}, which automatically  searches how to combine different components. For example, Su~\etal~\cite{su2021vision} searched patch size
and dimensions of linear projections and head number of attention modules to get an efficient vision transformer. Li~\etal~\cite{li2021bossnas} explored a self-supervised search strategy to get a hybrid architecture composing of both convolutional modules and self-attention modules.

\begin{figure}[t] 
	\centering
	\includegraphics[width=0.9\columnwidth]{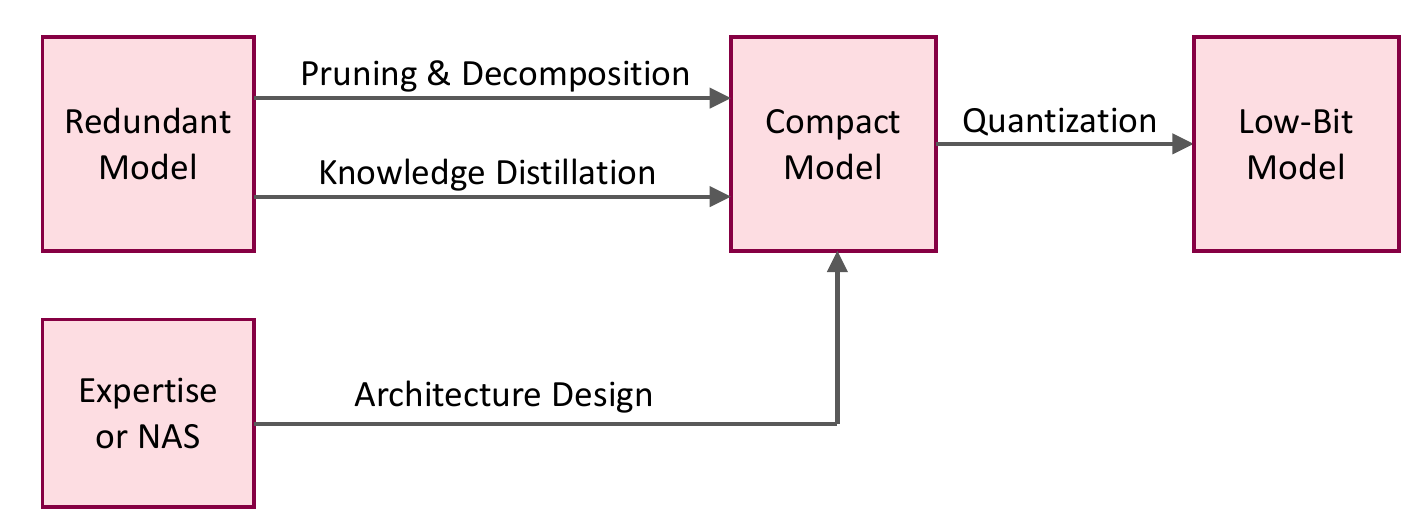}%
	\vspace{-0.5em}
	\caption{Different methods for compressing transformers.}
	\label{fig-relation}
	\vspace{-1.5em}
\end{figure}

The self-attention operation in transformer models calculates the dot product between  representations from different input tokens in a given sequence (patches in image recognition tasks~\cite{vit}), whose complexity is $O(N)$, where $N$ is the length of the sequence. Recently, there has been a targeted focus to reduce the complexity  to $O(N)$ in large methods so that transformer models can scale to long sequences~\cite{katharopoulos2020transformers, yun2020n,zaheer2020big}.   For example, Katharopoulos~\etal~\cite{katharopoulos2020transformers} approximated  self-attention as  a linear dot-product of kernel feature maps and revealed the relationship between tokens via RNNs. Zaheer~\etal~\cite{zaheer2020big} considered each token as a vertex in a graph and defined the inner product calculation between two tokens as an edge. Inspired by graph theories~\cite{spielman2011spectral,chung2002average}, various sparse graph are combined to approximate the dense graph in transformer models, and can achieve $O(N)$ complexity.

\textbf{Discussion.} The preceding methods take different approaches in how they attempt to identify redundancy in transformer models~(see Figure~\ref{fig-relation}). Pruning and decomposition  methods usually require  pre-defined models with redundancy. Specifically, pruning focuses on reducing the number of components (\eg, layers, heads) in transformer models while decomposition   represents an original matrix with multiple small matrices. Compact models also can be directly designed either manually (requiring sufficient expertise) or automatically (\eg, via NAS). The obtained compact models can be further represented with low-bits via quantization methods for  efficient deployment on resource-limited devices.

\section{Conclusions and Discussions}
Transformer is becoming a hot topic in the field of computer vision due to its competitive performance and tremendous potential compared with CNNs. To discover and utilize the power of transformer, as summarized in this survey, a number of methods have been proposed in recent years. These methods show excellent performance on a wide range of visual tasks, including backbone, high/mid-level vision, low-level vision, and video processing. Nevertheless, the potential of transformer for computer vision has not yet been fully explored, meaning that several challenges still need to be resolved. In this section, we discuss these challenges and provide insights on the future prospects.

\subsection{Challenges}
Although researchers have proposed many transformer-based models to tackle computer vision tasks, these works are only the first steps in this field and still have much room for improvement. For example, the transformer architecture in ViT~\cite{vit} follows the standard transformer for NLP~\cite{vaswani2017attention}, but an improved version specifically designed for CV remains to be explored. Moreover, it is necessary to apply transformer to more tasks other than those mentioned earlier.

The generalization and robustness of transformers for computer vision are also challenging. Compared with CNNs, pure transformers lack some inductive biases and rely heavily on massive datasets for large-scale training~\cite{vit}. Consequently, the quality of data has a significant influence on the generalization and robustness of transformers. Although ViT shows exceptional performance on downstream image classification tasks such as CIFAR~\cite{cifar} and VTAB~\cite{VTAB}, directly applying the ViT backbone on object detection has failed to achieve better results than CNNs~\cite{beal2020toward}. There is still a long way to go in order to better generalize pre-trained transformers on more generalized visual tasks. Practitioners concern the robustness of transformer (e.g. the vulnerability issue~\cite{cheng2019robust}). Although the robustness has been investigated in~\cite{zhang2020adversarial,mahmood2021robustness,mao2021rethinking}, it is still an open problem waiting to be solved.

Although numerous works have explained the use of transformers in NLP~\cite{serrano2019attention,wiegreffe2019attention}, it remains a challenging subject to clearly explain why transformer works well on visual tasks. The inductive biases, including translation equivariance and locality, are attributed to CNN’s success, but transformer lacks any inductive bias. The current literature usually analyzes the effect in an intuitive way~\cite{vit,chefer2020transformer}. For example, Dosovitskiy~\etal~\cite{vit} claim that large-scale training can surpass inductive bias. Position embeddings are added into image patches to retain positional information, which is important in computer vision tasks. Inspired by the heavy parameter usage in transformers, over-parameterization~\cite{livni2014computational,neyshabur2018towards} may be a potential point to the interpretability of vision transformers.

Last but not least, developing efficient transformer models for CV remains an open problem. Transformer models are usually huge and computationally expensive. For example, the base ViT model~\cite{vit} requires 18 billion FLOPs to process an image. In contrast, the lightweight CNN model GhostNet~\cite{ghostnet,tinynet} can achieve similar performance with only about 600 million FLOPs. Although several methods have been proposed to compress transformer, they remain highly complex. And these methods, which were originally designed for NLP, may not be suitable for CV. Consequently, efficient transformer models are urgently needed so that vision transformer can be deployed on resource-limited devices.

\subsection{Future Prospects}
In order to drive the development of vision transformers, we provide several potential directions for future study.

One direction is the effectiveness and the efficiency of transformers in computer vision. The goal is to develop highly effective and efficient vision transformers; specifically, transformers with high performance and low resource cost. The performance determines whether the model can be applied on real-world applications, while the resource cost influences the deployment on devices~~\cite{chen2014diannao,liao2019davinci}. The effectiveness is usually correlated with the efficiency, so determining how to achieve a better balance between them is a meaningful topic for future study.

Most of the existing vision transformer models are designed to handle only a single task. Many NLP models such as GPT-3~\cite{gpt3} have demonstrated how transformer can deal with multiple tasks in one model. IPT~\cite{chen2020pre} in the CV field is also able to process multiple low-level vision tasks, such as super-resolution, image denoising, and deraining. Perceiver~\cite{jaegle21a} and Perceiver IO~\cite{jaegle2021perceiver} are the pioneering models that can work on several domains including images, audio, multimodal, point clouds. We believe that more tasks can be involved in only one model. Unifying all visual tasks and even other tasks in one transformer (i.e., a grand unified model) is an exciting topic.

There have been various types of neural networks, such as CNN, RNN, and transformer. In the CV field, CNNs used to be the mainstream choice~\cite{resnet,efficientnet}, but now transformer is becoming popular. CNNs can capture inductive biases such as translation equivariance and locality, whereas ViT uses large-scale training to surpass inductive bias~\cite{vit}. From the evidence currently available~\cite{vit}, CNNs perform well on small datasets, whereas transformers perform better on large datasets. The question for the future is whether to use CNN or transformer.

By training with large datasets, transformers can achieve state-of-the-art performance on both NLP~\cite{gpt3,bert} and CV benchmarks~\cite{vit}. It is possible that neural networks need big data rather than inductive bias. In closing, we leave you with a question: Can transformer obtains satisfactory results with a very simple computational paradigm (\eg, with only fully connected layers) and massive data training?

\section*{Acknowledgement}
This research is partially supported by MindSpore (\url{https://mindspore.cn/}) and CANN (Compute Architecture for Neural Networks).

\appendix
\subsection*{A1. General Formulation of Self-attention}
\label{sec:formulation-sa}
The self-attention module~\cite{vaswani2017attention} for machine translation computes the responses at each position in a sequence by estimating attention scores to all positions and gathering the corresponding embeddings based on the scores accordingly. This can be viewed as a form of non-local filtering operations~\cite{non-local,nl05denoising}. We follow the convention~\cite{non-local} to formulate the self-attention module. Given an input signal (\eg, image, sequence, video and feature) $\mathbf X\in \mathbb{R}^{n\times d}$, where $n=h\times w$ (indicating the number of pixels in feature) and $d$ is the number of channels, the output signal is generated as:
\begin{equation}
	\mathbf y_i = \frac{1}{C(\mathbf x_i)}\sum_{\forall j}{f(\mathbf x_i, \mathbf x_j)g(\mathbf x_j)},
	\label{eq:self-attention}
\end{equation}
where $\mathbf x_i\in \mathbb{R}^{1\times d}$ and $\mathbf y_i\in \mathbb{R}^{1\times d}$ indicate the $i^{th}$ position (\eg, space, time and spacetime) of the input signal $\mathbf X$ and output signal $\mathbf Y$, respectively. Subscript $j$ is the index that enumerates all positions, and a pairwise function $f(\cdot)$ computes a representing relationship (such as affinity) between $i$ and all $j$. The function $g(\cdot)$ computes a representation of the input signal at position $j$, and the response is normalized by a factor $C(x_i)$.

Note that there are many choices for the pairwise function $f(\cdot)$. For example, a simple extension of the Gaussian function could be used to compute the similarity in an embedding space. As such, the function $f(\cdot)$ can be formulated as:
\begin{equation}
	f(\mathbf x_i, \mathbf x_j) = e^{\theta(\mathbf x_i)\phi(\mathbf x_j)^T}
\end{equation}
where $\theta(\cdot)$ and $\phi(\cdot)$ can be any embedding layers. If we consider the $\theta(\cdot), \phi(\cdot), g(\cdot)$ in the form of linear embedding: $\theta(\mathbf X)=\mathbf X\mathbf W_{\theta}$, $\phi(\mathbf X)=\mathbf X\mathbf W_{\phi}, g(\mathbf X)=\mathbf X\mathbf W_g$ where $\mathbf W_{\theta}\in \mathbb{R}^{d\times d_k}, \mathbf W_{\phi}\in \mathbb{R}^{d\times d_k}, \mathbf W_g\in \mathbb{R}^{d\times d_v}$, and set the normalization factor as $C(\mathbf x_i)=\sum_{\forall j}{f(\mathbf x_i, \mathbf x_j)}$, the Eq.~\ref{eq:self-attention} can be rewritten as:
\begin{equation}
	\mathbf y_i = \frac{e^{\mathbf x_iw_{\theta,i}\mathbf w_{\phi,j}^T\mathbf x_j^T}}{\sum_j{e^{\mathbf x_i\mathbf w_{\theta,i}\mathbf w_{\phi,j}^Tx_j^T}}}\mathbf x_j\mathbf w_{g,j},
\end{equation}
where $\mathbf w_{\theta,i}\in \mathbb{R}^{d\times 1}$ is the $i^{th}$ row of the weight matrix $W_{\theta}$. For a given index $i$, $\frac{1}{C(\mathbf x_i)}f(\mathbf x_i, \mathbf x_j)$ becomes the softmax output along the dimension $j$. The formulation can be further rewritten as:
\begin{equation}
	\mathbf Y = \mathrm{softmax}(\mathbf X\mathbf W_{\theta}\mathbf W_{\phi}^T\mathbf X)g(\mathbf X),
	\label{eq:self-attention-softmax}
\end{equation}
where $\mathbf Y\in \mathbb{R}^{n\times c}$ is the output signal of the same size as $\mathbf X$.
Compared with the query, key and value representations $\mathbf Q=\mathbf X\mathbf W_q, \mathbf K=\mathbf X\mathbf W_k, \mathbf V=\mathbf X\mathbf W_v$ from the translation module, once $\mathbf W_q=\mathbf W_{\theta}, \mathbf W_k=\mathbf W_{\phi}, \mathbf W_v=\mathbf W_g$, Eq.~\ref{eq:self-attention-softmax} can be formulated as:
\begin{equation}
	\mathbf Y = \mathrm{softmax}(\mathbf Q\mathbf K^T)\mathbf V = \mathrm{Attention}(\mathbf Q, \mathbf K, \mathbf V),
\end{equation}
The self-attention module~\cite{vaswani2017attention} proposed for machine translation is, to some extent, the same as the preceding non-local filtering operations proposed for computer vision.

Generally, the final output signal of the self-attention module for computer vision will be wrapped as:
\begin{equation}
	\mathbf Z = \mathbf Y\mathbf W^o + \mathbf X
\end{equation}
where $\mathbf Y$ is generated through Eq.~\ref{eq:self-attention-softmax}. If $\mathbf W^o$ is initialized as zero, this self-attention module can be inserted into any existing model without breaking its initial behavior.

\subsection*{A2. Revisiting Transformers for NLP}\label{Sec:Introduction}
Before transformer was developed, RNNs (~\eg, GRU~\cite{GRU} and LSTM~\cite{LSTM}) with added attention~\cite{bahdanau2014neural} empowered most of the state-of-the-art language models. However, RNNs require the information flow to be processed sequentially from the previous hidden states to the next one. This rules out the possibility of using acceleration and parallelization during training, and consequently hinders the potential of RNNs to process longer sequences or build larger models. In 2017, Vaswani~\etal~\cite{vaswani2017attention} proposed transformer, a novel encoder-decoder architecture built solely on multi-head self-attention mechanisms and feed-forward neural networks. Its purpose was to solve seq-to-seq natural language tasks (\eg, machine translation) easily by acquiring global dependencies. The subsequent success of transformer demonstrates that leveraging attention mechanisms alone can achieve performance comparable with attentive RNNs. Furthermore, the architecture of transformer lends itself to massively parallel computing, which enables training on larger datasets. This has given rise to the surge of large pre-trained models (PTMs) for natural language processing.

BERT~\cite{bert} and its variants (\eg, SpanBERT~\cite{Spanbert}, RoBERTa~\cite{roberta}) are a series of PTMs built on the multi-layer transformer encoder architecture. Two tasks are conducted on BookCorpus~\cite{bookcorpus} and English Wikipedia datasets at the pre-training stage of BERT: 1) Masked language modeling (MLM), which involves first randomly masking out some tokens in the input and then training the model to predict; 2) Next sentence prediction, which uses paired sentences as input and predicts whether the second sentence is the original one in the document. After pre-training, BERT can be fine-tuned by adding an output layer on a wide range of downstream tasks. More specifically, when performing sequence-level tasks (\eg, sentiment analysis), BERT uses the representation of the first token for classification; for token-level tasks (\eg, name entity recognition), all tokens are fed into the softmax layer for classification. At the time of its release, BERT achieved the state-of-the-art performance on 11 NLP tasks, setting a milestone in pre-trained language models. Generative Pre-trained Transformer models (\eg, GPT~\cite{GPT}, GPT-2~\cite{GPT2}) are another type of PTMs based on the transformer decoder architecture, which uses masked self-attention mechanisms. The main difference between the GPT series and BERT is the way in which pre-training is performed. Unlike BERT, GPT models are unidirectional language models pre-trained using Left-to-Right (LTR) language modeling. Furthermore, BERT learns the sentence separator ([SEP]) and classifier token ([CLS]) embeddings during pre-training, whereas these embeddings are involved in only the fine-tuning stage of GPT. Due to its unidirectional pre-training strategy, GPT achieves superior performance in many natural language generation tasks.  More recently, a massive transformer-based model called GPT-3, which has an astonishing 175 billion parameters, was developed~\cite{gpt3}. By pre-training on 45 TB of compressed plaintext data, GPT-3 can directly process different types of downstream natural language tasks without fine-tuning. As a result, it achieves strong performance on many NLP datasets, including both natural language understanding and generation. Since the introduction of transformer, many other models have been proposed in addition to the transformer-based PTMs mentioned earlier. We list a few representative models in Table~\ref{tbl:tf_based} for interested readers, but this is not the focus of our study.

\begin{table}[h]
	\centering
	\renewcommand\arraystretch{1.1}
	\setlength{\tabcolsep}{3pt}{
		\begin{threeparttable}[b]
			\caption{List of representative language models built on transformer. Transformer is the standard encoder-decoder architecture. Transformer Enc. and Dec. represent the encoder and decoder, respectively. Decoder uses mask self-attention to prevent attending to the future tokens. The data of the Table is from ~\cite{PTMS}.}
			\footnotesize
			\begin{tabular}{c|c|c|c}				
				\Xhline{1.2pt}
				Models & Architecture & \# of Params & Fine-tuning \\		
				\hline
				GPT~\cite{GPT} & Transformer Dec. & 117M & Yes \\
				GPT-2~\cite{GPT2}& Transformer Dec. & 117M-1542M & No \\
				GPT-3~\cite{gpt3}& Transformer Dec. & 125M-175B & No \\
				BERT~\cite{bert} & Transformer Enc. & 110M-340M & Yes \\
				RoBERTa~\cite{roberta} & Transformer Enc. & 355M & Yes \\
				\multirow{2}{*}{XLNet~\cite{XLNet}} & Two-Stream & \multirow{2}{*}{$\approx$ BERT} & \multirow{2}{*}{Yes} \\
				& Transformer Enc. &  &  \\
				ELECTRA~\cite{Electra} & Transformer Enc. & 335M & Yes \\
				UniLM~\cite{UniLim} & Transformer Enc. & 340M & Yes\\
				BART~\cite{BART}  & Transformer & 110\% of BERT & Yes\\
				T5~\cite{2020t5} & Transfomer & 220M-11B & Yes\\
				ERNIE (THU)~\cite{ERNIE} & Transformer Enc. & 114M & Yes\\
				KnowBERT~\cite{KnownBERT} & Transformer Enc. & 253M-523M & Yes\\
				\hline
			\end{tabular}
			\label{tbl:tf_based}
	\end{threeparttable}}
	\vspace{-0em}
\end{table}

Apart from the PTMs trained on large corpora for general NLP tasks, transformer-based models have also been applied in many other NLP-related domains and to multi-modal tasks. 

\noindent\textbf{BioNLP Domain.} Transformer-based models have outperformed many traditional biomedical methods. Some examples of such models include BioBERT~\cite{biobert}, which uses a transformer architecture for biomedical text mining tasks, and SciBERT~\cite{scibert}, which is developed by training transformer on 114M scientific articles (covering biomedical and computer science fields) with the aim of executing NLP tasks in the scientific domain more precisely. Another example is ClinicalBERT, proposed by Huang ~\etal~\cite{clinicalbert}. It utilizes transformer to develop and evaluate continuous representations of clinical notes. One of the side effects of this is that the attention map of ClinicalBERT can be used to explain predictions, thereby allowing high-quality connections between different medical contents to be discovered. 

The rapid development of transformer-based models on a variety of NLP-related tasks demonstrates its structural superiority and versatility, opening up the possibility that it will become a universal module applied in many AI fields other than just NLP. The following part of this survey focuses on the applications of transformer in a wide range of computer vision tasks that have emerged over the past two years.

\subsection*{A3. Self-attention for Computer Vision}\label{subsec:selfatt}
The preceding sections reviewed methods that use a transformer architecture for vision tasks. We can conclude that self-attention plays a pivotal role in transformer. The self-attention module can also be considered a building block of CNN architectures, which have low scaling properties concerning the large receptive fields. This building block is widely used on top of the networks to capture long-range interactions and enhance high-level semantic features for vision tasks. In this section, we delve deeply into the models based on self-attention designed for challenging tasks in computer vision. Such tasks include semantic segmentation, instance segmentation, object detection, keypoint detection, and depth estimation. Here we briefly summarize the existing applications using self-attention for computer vision.

\textbf{Image Classification.}
Trainable attention for classification consists of two main streams: hard attention~\cite{ba2014multiple,mnih2014recurrent,xu2015show} regarding the use of an image region, and soft attention~\cite{wang2017residual,jetley2018learn,han2018attribute,ramachandran2019stand} generating non-rigid feature maps. Ba~\etal~\cite{ba2014multiple} first proposed the term ``visual attention'' for image classification tasks, and used attention to select relevant regions and locations within the input image. This can also reduce the computational complexity of the proposed model regarding the size of the input image. For medical image classification, AG-CNN~\cite{guan2018diagnose} was proposed to crop a sub-region from a global image by the attention heat map. And instead of using hard attention and recalibrating the crop of feature maps, SENet~\cite{senet} was proposed to reweight the channel-wise responses of the convolutional features using soft self-attention. Jetley~\etal~\cite{jetley2018learn} used attention maps generated by corresponding estimators to reweight intermediate features in DNNs. In addition, Han~\etal~\cite{han2018attribute} utilized the attribute-aware attention to enhance the representation of CNNs.

\textbf{Semantic Segmentation.}
PSANet~\cite{psanet}, OCNet~\cite{ocnet}, DANet~\cite{danet} and CFNet~\cite{cfnet} are the pioneering works to propose using the self-attention module in semantic segmentation tasks. These works consider and augment the relationship and similarity~\cite{acfnet,emanet,apcnet,oktay2018attention,wang2020self,li2019global} between the contextual pixels. DANet~\cite{danet} simultaneously leverages the self-attention module on spatial and channel dimensions, whereas $A^2$Net~\cite{a2net} groups the pixels into a set of regions, and then augments the pixel representations by aggregating the region representations with the generated attention weights. DGCNet~\cite{zhang2019dual} employs a dual graph CNN to model coordinate space similarity and feature space similarity in a single framework. To improve the efficiency of the self-attention module for semantic segmentation, several works~\cite{cgnl,ccnet,isanet,li2018beyond,kumaar2020cabinet} have been proposed, aiming to alleviate the huge amount of parameters brought by calculating pixel similarities. For example, CGNL~\cite{cgnl} applies the Taylor series of the RBF kernel function to approximate the pixel similarities. CCNet~\cite{ccnet} approximates the original self-attention scheme via two consecutive criss-cross attention modules. In addition, ISSA~\cite{isanet} factorizes the dense affinity matrix as the product of two sparse affinity matrices. There are other related works using attention based graph reasoning modules~\cite{liang2018symbolic,chen2019graph,li2018beyond} to enhance both the local and global representations.

\textbf{Object Detection.}
Ramachandran~\etal~\cite{ramachandran2019stand} proposes an attention-based layer and swapped the conventional convolution layers to build a fully attentional detector that outperforms the typical RetinaNet~\cite{retinanet} on COCO benchmark~\cite{coco}. GCNet~\cite{gcnet} assumes that the global contexts modeled by non-local operations are almost the same for different query positions within an image, and unifies the simplified formulation and SENet~\cite{senet} into a general framework for global context modeling~\cite{li2020object,hsieh2019one,fan2020few,perreault2020spotnet}.
Vo~\etal~\cite{vo2020bidirectional} designs a bidirectional operation to gather and distribute information from a query position to all possible positions. Zhang~\etal~\cite{zhang2020feature} suggests that previous methods fail to interact with cross-scale features, and proposes Feature Pyramid Transformer, based on the self-attention module, to fully exploit interactions across both space and scales.

Conventional detection methods usually exploit a single visual representation (\eg, bounding box and corner point) for predicting the final results. Hu~\etal~\cite{relationnet} proposes a relation module based on self-attention to process a set of objects simultaneously through interaction between their appearance features. Cheng~\etal~\cite{relationnet++} proposes RelationNet++ with the bridging visual representations (BVR) module to combine different heterogeneous representations into a single one similar to that in the self-attention module. Specifically, the master representation is treated as the query input and the auxiliary representations are regarded as the key input. The enhanced feature can therefore bridge the information from auxiliary representations and benefit final detection results.

\textbf{Other Vision Tasks.}
Zhang~\etal~\cite{zhang2020learning} proposes a resolution-wise attention module to learn enhanced feature maps when training multi-resolution networks to obtain accurate human keypoint locations for pose estimation task. Furthermore, Chang~\etal~\cite{chang2019same} uses an attention-mechanism based feature fusion block to improve the accuracy of the human keypoint detection model.

To explore more generalized contextual information for improving the self-supervised monocular trained depth estimation, Johnston~\etal~\cite{johnston2020self} directly leverages self-attention module. Chen~\etal~\cite{chen2019attention} also proposes an attention-based aggregation network to capture context information that differs in diverse scenes for depth estimation. And Aich~\etal~\cite{aich2020bidirectional} proposes bidirectional attention modules that utilize the forward and backward attention operations for better results of monocular depth estimation.

\ifCLASSOPTIONcaptionsoff
  \newpage
\fi

\renewcommand\refname{References}

{\small
\bibliographystyle{unsrt2authabbrvpp}
\bibliography{ref}
}

\end{document}